\documentclass[manuscript,screen,nonacm]{acmart}

\AtBeginDocument{%
  }

\setcopyright{acmlicensed}
\copyrightyear{2018}
\acmYear{2018}
\acmDOI{XXXXXXX.XXXXXXX}

\acmBooktitle{Woodstock '18: ACM Symposium on Neural Gaze Detection,
 June 03--05, 2018, Woodstock, NY}
\acmISBN{978-1-4503-XXXX-X/18/06}

\usepackage{array}
\usepackage[caption=false,font=normalsize,labelfont=sf,textfont=sf]{subfig}
\usepackage{wrapfig}
\usepackage{tabularx}
\usepackage{textcomp}
\usepackage{stfloats}
\usepackage{url}
\usepackage{verbatim}

\usepackage{graphicx}
\usepackage{multirow}
\usepackage{booktabs}
\usepackage{tabularx}
\usepackage{amsfonts}%

\usepackage{amsthm}%
\usepackage{mathrsfs}%
\usepackage{xcolor}%
\usepackage{textcomp}%
\usepackage{manyfoot}%
\usepackage{booktabs}%
\usepackage{algorithm}%
\usepackage{algorithmicx}%
\usepackage{algpseudocode}
\usepackage{multicol}
\usepackage{caption}
\usepackage{listings}%
\usepackage{longtable}
\usepackage{ltxtable}
\usepackage{array}
\usepackage{tabularray}
\usepackage{bbding}
\usepackage{colortbl}
\usepackage{xspace}
\usepackage{url}
\usepackage{arydshln}

\definecolor{dino}{RGB}{249,231,227}
\definecolor{image}{RGB}{173,216,230}
\definecolor{video}{RGB}{135,206,235}
\definecolor{pointcloud}{RGB}{0,191,255}
\definecolor{multimodal}{RGB}{30,144,255}
\definecolor{myblue}{RGB}{112, 219, 219}
\definecolor{commentcolor}{RGB}{34, 139, 34} 
\definecolor{codecolor}{RGB}{0, 0, 0} 
\definecolor{algocolor}{RGB}{128, 0, 0} 
\definecolor{SSMd}{RGB}{240, 201, 207} %

\usepackage[edges]{forest}
\definecolor{lightcoral}{rgb}{0.94, 0.5, 0.5}
\definecolor{lightgreen}{rgb}{0.56, 0.93, 0.56}

\definecolor{brightlavender}{rgb}{0.75, 0.58, 0.89}
\definecolor{capri}{rgb}{0.0, 0.75, 1.0}
\definecolor{carminepink}{rgb}{0.92, 0.3, 0.26}
\definecolor{celadon}{rgb}{0.67, 0.88, 0.69}
\definecolor{darkpastelgreen}{rgb}{0.01, 0.75, 0.24}

\definecolor{pastelblue}{rgb}{0.68, 0.78, 0.81}
\definecolor{mintgreen}{rgb}{0.6, 0.98, 0.6}
\definecolor{lavender}{rgb}{0.71, 0.49, 0.86}
\definecolor{peach}{rgb}{1.0, 0.9, 0.71}
\definecolor{coral}{rgb}{1.0, 0.5, 0.31}
\definecolor{mauve}{rgb}{0.88, 0.69, 1.0}
\definecolor{lemonyellow}{rgb}{1.0, 0.96, 0.4}

\definecolor{hidden-draw}{RGB}{205, 44, 36}
\definecolor{hidden-blue}{RGB}{194,232,247}
\definecolor{hidden-orange}{RGB}{243,202,120}
\definecolor{hidden-yellow}{RGB}{242,244,193}
\definecolor{tree-level-1}{RGB}{245,20,85}
\definecolor{tree-level-2}{RGB}{246,86,118}
\definecolor{tree-level-3}{RGB}{248,177,193}
\definecolor{tree-leaf}{RGB}{176,230,198}

\definecolor{Self}{RGB}{255,0,128}
\definecolor{Ensemble}{RGB}{0,127,255}
\definecolor{Iterative}{RGB}{153,51,255}

\definecolor{exemplar1}{RGB}{136,98,148}
\definecolor{exemplar2}{RGB}{148,210,242}
\definecolor{knowledge1}{RGB}{249,219,152}
\definecolor{knowledge2}{RGB}{255,245,220}

\definecolor{lighttealblue}{RGB}{41, 157, 143}
\definecolor{lightplum}{RGB}{233, 196, 106}
\definecolor{harvestgold}{RGB}{216, 118, 89}
\definecolor{lightblue}{RGB}{186, 204, 217}
\definecolor{lightgreen}{RGB}{173, 213, 162}

\definecolor{top1}{RGB}{255, 102, 102}
\definecolor{top2}{RGB}{102, 153, 255}
\definecolor{top3}{RGB}{102, 255, 178}

\usepackage{enumitem}
\usepackage{caption}

\algnewcommand{\LineComment}[1]{\State \textcolor{commentcolor}{\# #1}}

\newcommand{\eg}{\emph{e.g.}}

\begin{document}

\title{Visual Mamba: A Survey and New Outlooks}


\author{Rui Xu}
\orcid{0000-0002-0737-5214}
\affiliation{%
  \institution{The Hong Kong University of Science and Technology}
  \city{Hong Kong}
  \country{China}}
\email{cseruixu@ust.hk}

\author{Shu Yang}
\orcid{0000-0002-1761-9286}
\affiliation{%
  \institution{The Hong Kong University of Science and Technology}
  \city{Hong Kong}
  \country{China}}
\email{syangcw@connect.ust.hk}

\author{Yihui Wang}
\orcid{0009-0002-7606-9816}
\affiliation{%
  \institution{The Hong Kong University of Science and Technology}
  \city{Hong Kong}
  \country{China}}
\email{ywangrm@connect.ust.hk}

\author{Yu Cai}
\orcid{0000-0002-7749-8463}
\affiliation{%
  \institution{The Hong Kong University of Science and Technology}
  \city{Hong Kong}
  \country{China}}
\email{yu.cai@connect.ust.hk}

\author{Bo Du}
\orcid{0000-0002-0059-8458}
\affiliation{%
  \institution{Wuhan University}
  \city{Wuhan}
  \country{China}}
\email{dubo@whu.edu.cn}

\author{Hao Chen}\authornote{Corresponding Author}
\orcid{0000-0002-8400-3780}
\affiliation{%
  \institution{The Hong Kong University of Science and Technology}
  \city{Hong Kong}
  \country{China}}
\email{jhc@cse.ust.hk}


\begin{abstract}
Mamba, a recent selective structured state space model, excels in long sequence modeling, which is vital in the large model era. Long sequence modeling poses significant challenges, including capturing long-range dependencies within the data and handling the computational demands caused by their extensive length. Mamba addresses these challenges by overcoming the local perception limitations of convolutional neural networks and the quadratic computational complexity of Transformers. Given its advantages over these mainstream foundation architectures, Mamba exhibits great potential to be a visual foundation architecture. Since January 2024, Mamba has been actively applied to diverse computer vision tasks, yielding numerous contributions. To help keep pace with the rapid advancements, this paper reviews visual Mamba approaches, analyzing nearly 300 papers. This paper begins by delineating the formulation of the original Mamba model. Subsequently, it delves into representative backbone networks, and applications categorized using different modalities, including image, video, point cloud, and multi-modal data. Particularly, we identify scanning techniques as critical for adapting Mamba to vision tasks, and decouple these scanning techniques to clarify their functionality and enhance their flexibility across various applications. Finally, we discuss the challenges and future directions, providing insights into new outlooks in this fast evolving area. A comprehensive list of visual Mamba models reviewed in this work is available at \url{https://github.com/Ruixxxx/Awesome-Vision-Mamba-Models}.
\end{abstract}



\renewcommand{\keywordsname}{Keywords}
\keywords{Mamba, State Space Model, Computer Vision, Application}


\maketitle

\section{Introduction}
Artificial intelligence technologies, especially deep learning, have revolutionized numerous application fields. In the field of computer vision (CV), convolutional neural networks (CNNs) utilize local receptive fields and shared weights to process visual data, capitalizing on inductive biases such as locality and spatial invariance \cite{nips12/cnn,iclr15/cnn2,cvpr16/resnet}. Despite their efficient linear computational complexity with respect to image resolution and proficiency in modeling local patterns, CNNs have restricted receptive fields. This restriction limits their capability to capture larger spatial contexts, which is essential for comprehensively understanding scenes or complex spatial relations in tasks that demand global information. In recent years, Vision Transformers (ViTs) \cite{iclr21/vit}, which utilize a self-attention mechanism \cite{nips17/attention} to process sequences of image patches, have demonstrated remarkable modeling capabilities across various visual tasks \cite{iccv21/swinvit}. Self-attention enables ViTs to capture long-range dependencies within images, providing a significant advantage over traditional CNNs that rely on local receptive fields. This capability allows ViTs to exhibit robust performance on varied datasets and scale effectively to large model sizes. 
However, the self-attention mechanism involves a quadratic computational cost to the number of patches, which limits the scalability of ViTs. The CV domain has long been dominated by CNNs and ViTs, each with their respective strengths and inherent limitations. To overcome their limitations, researchers have devoted significant effort to improving these models. Recently, structured state space models \cite{iclr22/s4,iclr23/s4_legs} have garnered considerable attention due to their computational efficiency and principled capability in modeling long-range dependencies \cite{nips20/hippo}. 

\begin{wrapfigure}{r}{0.5\textwidth}
  \centering
  \includegraphics[width=0.48\textwidth]{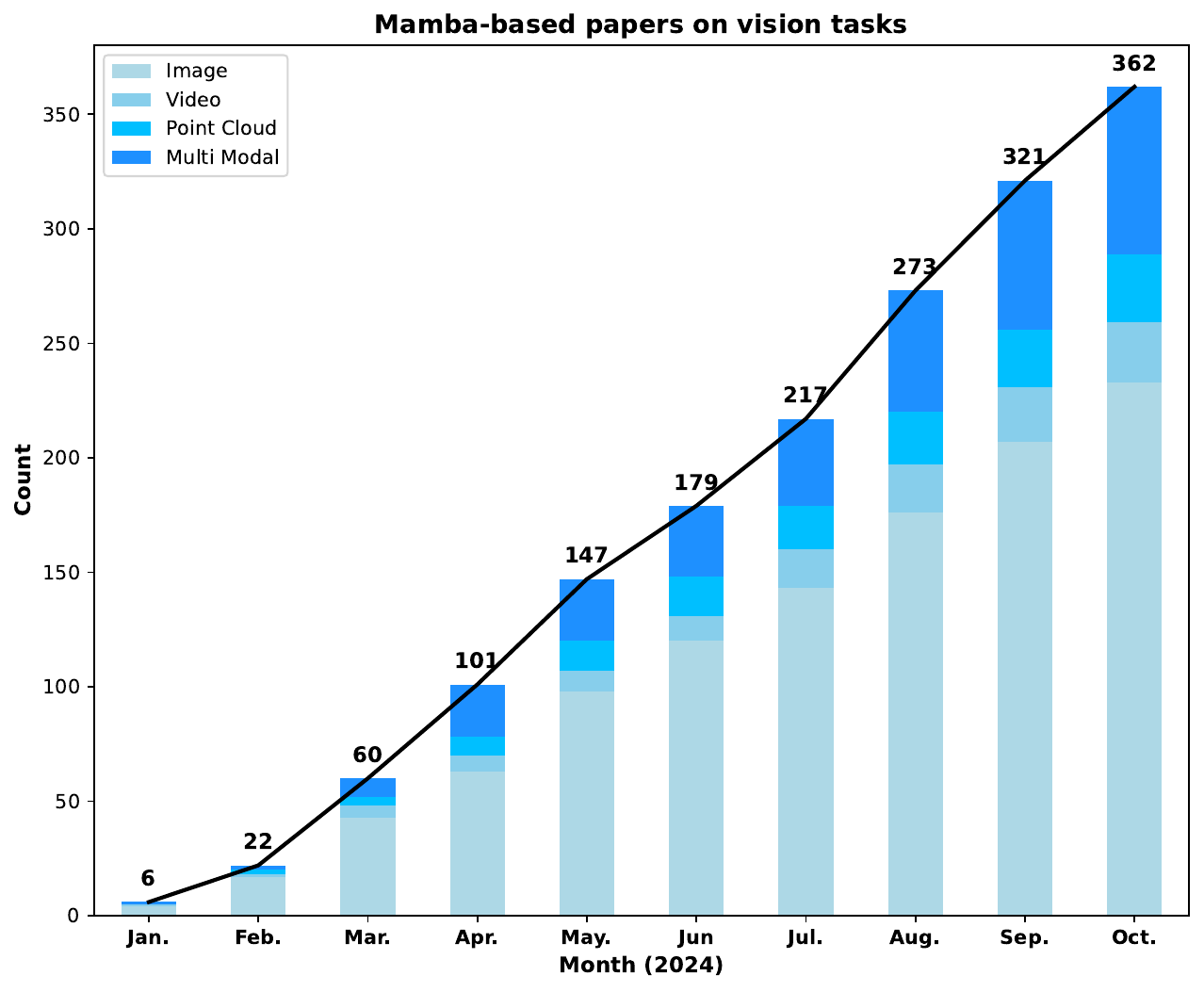}
  \caption{The statistics of Mamba-based papers released to date on vision tasks, spanning different modalities including \textcolor{image}{Image}, \textcolor{video}{Video}, \textcolor{pointcloud}{Point Cloud}, and \textcolor{multimodal}{Multi-Modal}.
  }
  \Description{}
  \label{fig:count}
\end{wrapfigure}

The state space model is a concept that is widely adopted in various disciplines. Its core idea is connecting the input and output sequences using a latent state. It takes different forms in different disciplines, such as Markov decision process in reinforcement learning \cite{iclr20/markov}, dynamic causal modeling in computational neuroscience \cite{neuroimage03/dcm} and Kalman filters in controls \cite{kalman1960}. Recently, the state space model (SSM) has been introduced to deep learning for sequence modeling and its parameters or mappings are learned by gradient descent \cite{nips21/LSSL}. SSM is essentially a type of sequence transformation and can be incorporated into deep neural networks. It conceptually unifies the strengths of former paradigms for sequence model design, including continuous-time models (CTMs), recurrent neural networks (RNNs), and CNNs. Among these, although CNNs are primarily used for processing spatial data and are not sequential, they can be adapted for sequence modeling by leveraging their abilities to encode local contexts and facilitate parallelizable computations. However, SSMs have not been widely used in practice due to their extensive computational and memory requirements coming from the state representation. This situation has changed with the advent of the structured SSM (S4), which addresses these limitations through reparameterizing the state matrices \cite{iclr22/s4}. Since then, a series of SSM transformations and neural network architectures incorporating SSM layers emerge \cite{iclr22/s4,nips22/s4d,nips22/dss,icml23/lru,iclr23/s5,iclr23/s4_legs,lclr23/liquid_s4, iclr23/h3}. However, SSMs' constant sequence transitions restrict their context-based reasoning ability, which is important for the efficacy of models like Transformer \cite{nips17/attention}. In Mamba-1 \cite{arxiv23/mamba}, the authors propose to address this by integrating a selection mechanism into the SSM, thus enabling the SSM to selectively propagate or forget information along the sequence or scan path based on the current token. Furthermore, to efficiently compute these selective SSMs, the authors develop a hardware-aware algorithm. 
In Mamba-2~\cite{arxiv24/mamba-2}, the authors establish connections between the SSMs and variants of attention, and refine the selective SSM into a state space duality (SSD) algorithm to further improve computational efficiency. With the modeling power akin to Transformers and linear scalability in sequence length, Mamba becomes a promising foundation architecture for sequence modeling, which is crucial in the large model era. 

Due to the growing adoption of techniques from sequence modeling or natural language processing (NLP) into CV, there is a rapid application of Mamba to CV tasks \cite{zhang2024surveyvisualmamba,liu2024visionmambacomprehensivesurvey,heidari2024computationefficienteracomprehensivesurvey,zou2024venturingunchartedwatersnavigation}. VMamba \cite{neurips24/vmamba}, an early representative visual Mamba model, unfolds image patches into sequences along the horizontal and vertical dimensions of the image, and performs bi-directional scanning along these two directions. Another visual Mamba model Vim \cite{icml24/vim} leverages position embeddings to incorporate spatial information, inspired by ViT \cite{iclr21/vit}. It also uses bi-directional SSM for handling the non-causal image sequences. Similarly, several other notable studies \cite{eccv24/Mamba-ND,bmvc24/plainmamba,arxiv24/simba,arxiv24/localmamba,arxiv24/effvmamba} delve into the exploration of visual backbone networks, consistently achieving competitive performance across classification, detection, and segmentation tasks. 
To highlight the balance between efficiency and effectiveness in visual Mamba models, Fig.~\ref{fig:cls_compare} offers a graphical representation contrasting their performance with computational complexity. In addition to these efforts, Mamba has been applied across a diverse range of vision modalities and their respective applications, encompassing image processing, video analysis, point cloud processing, and multi-modal scenarios. Endowed with the aforementioned modeling capability and linear scalability, Mamba stands out as a promising foundation architecture for CV tasks. The growing interest among researchers in applying Mamba to various vision tasks is reflected in the increasing number of studies dedicated to this exploration, as plotted in Fig.~\ref{fig:count}. 

In the rapidly evolving field of CV, Mamba has emerged as a significant advancement. Keeping pace with the latest research is critical for the community. Therefore, this paper aims to provide a comprehensive review of the applications of Mamba in vision tasks, shedding light on both its foundational elements and diverse applications across various modalities. Our contributions are summarized as follows: 

\begin{enumerate}[label=\arabic{enumi}.]
    \item \textbf{Formulation of Mamba} (Section~\ref{sec:form_of_mamba}): We provide an introductory overview of the operational principles of the Mamba \cite{arxiv23/mamba,arxiv24/mamba-2} and highlight its distinctions from traditional state space models.
    \item \textbf{Backbone Networks} (Section~\ref{sec:backbone}): We provide a detailed examination of several representative visual Mamba backbone networks. This analysis aims to elucidate the core principles and innovations that underpin the visual Mamba framework. 
    \item \textbf{Applications} (Section~\ref{sec:applications}): We categorize applications of Mamba by different modalities, such as image, video, point cloud, and multi-modal data. Each category is explored in depth to highlight how the Mamba architecture adapts to and benefits individual modalities. 
    \item \textbf{Challenges and Future Directions} (Section~\ref{sec:challenges_and_future}): We examine the challenges of visual Mamba models, focusing on their scalability, causality, in-context learning, and trustworthiness. More importantly, we outline prospective directions for visual Mamba models, providing new outlooks into exploring their untapped potential.
\end{enumerate}

\section{Formulation of Mamba}
\label{sec:form_of_mamba}
Mamba \cite{arxiv23/mamba} is a recent sequence model aiming at improving the context-based reasoning ability of state space model (SSM) by simply making its parameters to be functions of the input. The SSM here especially refers to the sequence transformation used in the structured state space sequence model (S4) \cite{iclr22/s4}, which can be incorporated into deep neural networks. 
Mamba-2~\cite{icml24/mamba2} further provides a theoretical framework that connects structured SSMs and variants of attention, thereby enabling the transfer of algorithmic and system optimizations originally developed for Transformers to SSMs. In the following, we elaborate on the core concepts of Mamba-1 and Mamba-2.

\subsection{State Space Model}
The SSM transformation in S4 \cite{iclr22/s4} originates from the classical state space model, which maps a 1D input signal $x(t) \in \mathbb{R}$ to a 1D output signal $y(t) \in \mathbb{R}$ through an N-D latent state $h(t) \in \mathbb{R}^{N}$. This transformation is mathematically formulated as linear ordinary differential equations (ODEs):
\begin{equation}
\begin{aligned}
h^{\prime}(t) = \mathbf{A}h(t) + \mathbf{B}x(t), \quad y(t) = \mathbf{C}h(t),
\label{eq:ssm}
\end{aligned}
\end{equation}
where $\mathbf{A} \in \mathbb{R}^{N \times N}$ is the evolution parameter and $\mathbf{B} \in \mathbb{R}^{N \times 1}, \mathbf{C} \in \mathbb{R}^{1 \times N}$ are the projection parameters of neural networks in deep learning. The term $h^{\prime}(t)$ denotes the derivative of $h(t)$ with respect to time $t$. 

To deal with the discrete input sequence $\boldsymbol{x} = (x_0, x_1, ...)  \in \mathbb{R}^{L}$, various discretization rules can be used to discretize the parameters in Eq.~\eqref{eq:ssm} using a step size $\Delta$, which can be seen as the resolution of the continuous input $x(t)$. Following previous work \cite{tustin1947method}, S4 \cite{iclr22/s4} discretizes these parameters using the bilinear method. Following \cite{nips22/dss}, Mamba \cite{arxiv23/mamba} employs the zero-order hold (ZOH) assumption\footnote{ZOH assumes that the sample value of $x$ remains constant over each sampling interval $\Delta$. For more details on ZOH, refer to \url{https://en.wikipedia.org/wiki/Zero-order_hold}} to solve the ODEs. Specifically, this results in the conversion of the continuous parameters $\mathbf{A}, \mathbf{B}$ into their discrete counterparts $\overline{\mathbf{A}}, \overline{\mathbf{B}}$ as follows:
\begin{equation}
\begin{aligned}
\overline{\mathbf{A}} = \exp(\Delta\mathbf{A}), \quad \overline{\mathbf{B}} = (\Delta\mathbf{A})^{-1}(\exp(\Delta\mathbf{A}) - \mathbf{I}) \cdot \Delta\mathbf{B}.
\end{aligned}
\end{equation}

After discretizing $\mathbf{A}, \mathbf{B}$ to $\overline{\mathbf{A}}, \overline{\mathbf{B}}$, the Eq.~\eqref{eq:ssm} can be reformulated as:
\begin{equation}
\begin{aligned}
h_t = \overline{\mathbf{A}}h_{t-1} + \overline{\mathbf{B}}x_t, \quad y_t = \mathbf{C}h_t.
\label{eq:ssm_d}
\end{aligned}
\end{equation}

Eq.~\eqref{eq:ssm_d} represents a sequence-to-sequence mapping from $x_t$ to $y_t$. This configuration allows the discretized SSM to be computed as RNN. However, due to its sequential nature, this discretized recurrent SSM is impractical for training. 

For efficient parallelizable training, 
this recursive process can be reformulated and computed as a convolution \cite{iclr22/s4}:
\begin{equation}
\begin{aligned}
\overline{\mathbf{K}} = (\mathbf{C}\overline{\mathbf{B}}, \mathbf{C}\overline{\mathbf{AB}}, \ldots, \mathbf{C}\overline{\mathbf{A}}^{L-1}\overline{\mathbf{B}}), \quad
\boldsymbol{y} = \boldsymbol{x} * \overline{\mathbf{K}},
\label{eq:ssm_c}
\end{aligned}
\end{equation}
where $L$ denotes the length of the input sequence $\boldsymbol{x}$ and $*$ represents the convolution operation. The vector $\overline{\mathbf{K}} \in \mathbb{R}^{L}$ is the SSM convolution kernel, enabling the simultaneous synthesis of outputs across the sequence. Given $\overline{\mathbf{K}}$, the convolution operation in Eq.~\eqref{eq:ssm_c} can be efficiently computed using the fast Fourier transforms (FFTs).

\subsection{Mamba-1: Selective SSM}
As seen, the parameters in SSM indicated by either Eq.~\eqref{eq:ssm}, Eq.~\eqref{eq:ssm_d} or Eq.~\eqref{eq:ssm_c} remain invariant with respect to the input or temporal dynamics. Mamba \cite{arxiv23/mamba} identifies this linear time-invariant (LTI) property as a fundamental limitation of SSM when it comes to context-based reasoning. To address this issue, Mamba incorporates a selection mechanism. 
The selection mechanism is implemented by simply configuring the parameters of SSM as functions of the input, thus achieving input-dependent interactions along the sequence. Specifically, parameters $\mathbf{B}, \mathbf{C}, \Delta$ are dependent on the input sequence $\boldsymbol{x}$:
\begin{equation}
\begin{aligned}
    \mathbf{B} = s_{B}(\boldsymbol{x}), \quad \mathbf{C} = s_{C}(\boldsymbol{x}), \quad \Delta = \tau_{\Delta}(Parameter + s_{\Delta}(\boldsymbol{x})),
\end{aligned}
\end{equation}
\sloppy
where $s_{B}(\boldsymbol{x})=Linear_{N}(\boldsymbol{x})$ and $s_{C}(\boldsymbol{x})=Linear_{N}(\boldsymbol{x})$ both project the input into a dimension $N$. $s_{\Delta}(\boldsymbol{x})=Broadcast_{D}(Linear_{1}(\boldsymbol{x}))$ first projects the input to dimension 1, and then broadcasts it to dimension $D$. $\tau_{\Delta}$ is the $softplus$ function. Here we present the complete shape of $\boldsymbol{x} \in \mathbb{R}^{B \times L \times D}$, where $B$ denotes the batch size, $L$ is the sequence length, and $D$ is the number of channels. Accordingly, the parameters $\mathbf{B}$ and $\mathbf{C}$ are each shaped as $\mathbb{R}^{B \times L \times N}$, and $\Delta$ is shaped as $\mathbb{R}^{B \times L \times D}$. 

The resulting selective SSM cannot be computed as either RNN or convolution. Mamba \cite{arxiv23/mamba} employs a hardware-aware algorithm to efficiently compute the selective SSM. The hardware-aware algorithm leverages three classical techniques: parallel scan, kernel fusion, and recomputation.

\subsection{Mamba-2: State Space Duality}
To leverage the collective advancements made by the community on Transformers, 
Mamba-2~\cite{arxiv24/mamba-2} first reformulates the SSM transformation through induction as:
\begin{equation}
    y_t = \sum_{s=0}^{t} \mathbf{C}_t^\top \mathbf{A}_{t:s}^\times \mathbf{B}_s x_s,
\end{equation}
where $\mathbf{A}_{t:s}^\times = \mathbf{A}_{t} \times \cdots \times \mathbf{A}_{s+1}, s<t$. The matrix transformation form of SSM can then be derived as:

\begin{equation}
    \boldsymbol{y} = \text{SSM}(\mathbf{A}, \mathbf{B}, \mathbf{C})(\boldsymbol{x}) = \mathbf{M}\boldsymbol{x},
\end{equation}
where $\mathbf{M}_{ji} = \mathbf{C}_j^\top \mathbf{A}_j \cdots \mathbf{A}_{i+1} \mathbf{B}_i$. Consequently, SSM also has a quadratic algorithm besides the linear algorithm. 

Mamba-2 considers a special case of structured SSM when $\mathbf{A}_{j}$ is a scalar. Then the matrix can be rearranged as:
\begin{equation}
    \mathbf{M}_{ji} = \mathbf{A}_{j:i} \cdot (\mathbf{C}_j^\top \mathbf{B}_i).
\end{equation}

This can be vectorized into:
\begin{equation}
    \mathbf{M} = \mathbf{L} \circ (\mathbf{C} \mathbf{B}^\top),
\end{equation}
which is consistent with the original definition of masked kernel attention. To sum up, SSM with a scalar-identity structure in the $\mathbf{A}$ matrix, and structured masked attention with a 1-semiseparable structure in the $\mathbf{L}$ mask are dual to each other, exhibiting identical linear and quadratic forms.

Mamba-2 introduces an SSD algorithm based on block decompositions of $\mathbf{M}$, leveraging both the linear SSM recurrence and its quadratic dual form to improve computational and memory efficiency while utilizing matrix multiplication units on modern hardware. This SSD algorithm achieves a $2$-$8 \times$ speedup over the hardware-aware Mamba-1 implementation.

\subsection{Mamba Architecture}
\begin{figure}[!t]
  \centering
  \includegraphics[width=0.8\textwidth]{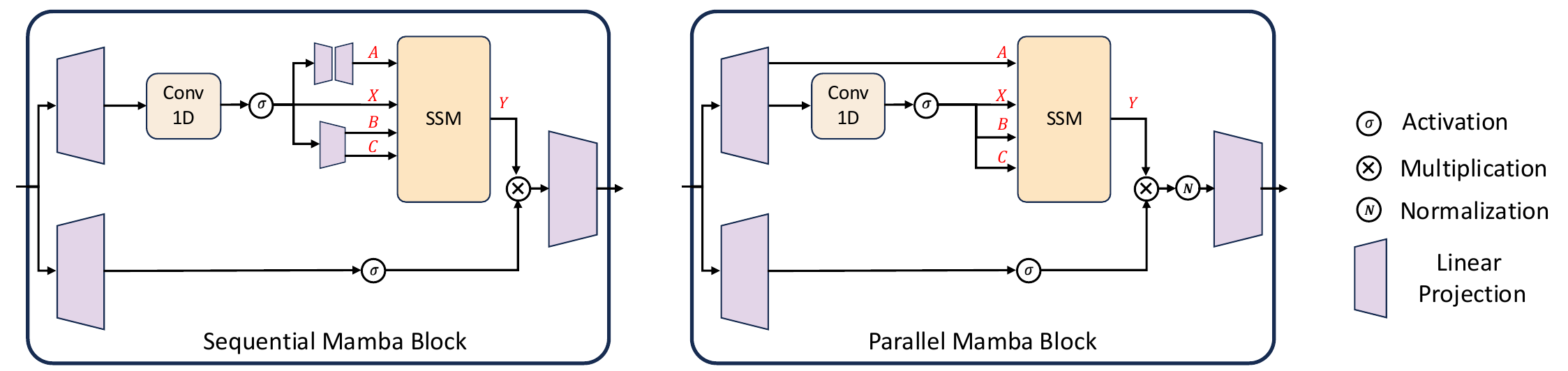}
  \caption{Mamba-1 block: sequential generation of SSM parameters vs. Mamba-2 block: parallel generation of SSM parameters.}
  \Description{}
  \label{fig:mamba_layer}
\end{figure}

Mamba-1 is a simplified SSM architecture, drawing insights from the H3 architecture \cite{iclr23/h3}, which serves as the basis of many SSM architectures. Specifically, the H3 \cite{iclr23/h3} block integrates two discrete SSMs to explicitly simulate linear attention. The commonly used SSM architectures alternately stack the H3 block \cite{iclr23/h3} and the MLP block prevalent in modern neural networks. Unlike them, Mamba-1 integrates these two blocks to construct the simplified Mamba-1 block, as illustrated in Fig.~\ref{fig:mamba_layer}. The Mamba-1 block can be viewed from two distinct perspectives. Firstly, it replaces the first multiplicative gate in the linear attention-like or H3 \cite{iclr23/h3} block with an activation function. Secondly, it incorporates the SSM transformation into the primary pathway of the MLP block. The overall architecture of Mamba-1 consists of repeated Mamba-1 blocks interleaved with standard normalization layers and residual connections.

Different from Mamba-1, which maps from $\boldsymbol{x}$ to $\boldsymbol{y}$ under the SSM point of view and sequentially computes $\mathbf{A}, \mathbf{B}, \mathbf{C}$ after $\boldsymbol{x}$, Mamba-2 produces $\mathbf{A}, \boldsymbol{x}, \mathbf{B}, \mathbf{C}$ in parallel as the SSD layer is regarded as a mapping from $\mathbf{A}, \boldsymbol{x}, \mathbf{B}, \mathbf{C}$ to $\boldsymbol{y}$. Thus the Mamba-2 block simplifies the Mamba-1 block by removing sequential linear projections; namely, SSM parameters $\mathbf{A}, \mathbf{B}, \mathbf{C}$ are generated at the beginning of the block, independent of the SSM input $\boldsymbol{x}$, as illustrated in Fig.~\ref{fig:mamba_layer}. Additionally, a normalization layer as in NormFormer~\cite{arxiv2021/NormFormer} is added after the multiplicative gate branch to enhance stability.

Mamba inherits the linear scalability in sequence length of the state space models, and also realizes the modeling ability of Transformers. Mamba exhibits the significant advantages of two primary types of foundation architectures in CV, i.e. CNNs and Transformers, making it a promising foundation architecture for CV. In contrast to Transformers, which rely on explicitly storing the entire context for context-based reasoning, Mamba utilizes a selection mechanism, which selectively propagates or filters information along the sequence based on the current token. This ensures the modeling process is influenced solely by the past and current inputs along the sequence, thereby adhering to the principles of causality. Therefore, adapting the inherent 1D and causal characteristics of this selection mechanism to align with the nature of visual data has become a primary focus for researchers applying Mamba to CV.

\section{Visual Mamba Backbone Networks}
\label{sec:backbone}
This section reviews representative visual Mamba backbone networks, categorizing them by key architectural advancements and optimization strategies. A comparative analysis of their performance is also presented.

\begin{figure}[!t]
  \centering
  \includegraphics[width=0.8\textwidth]{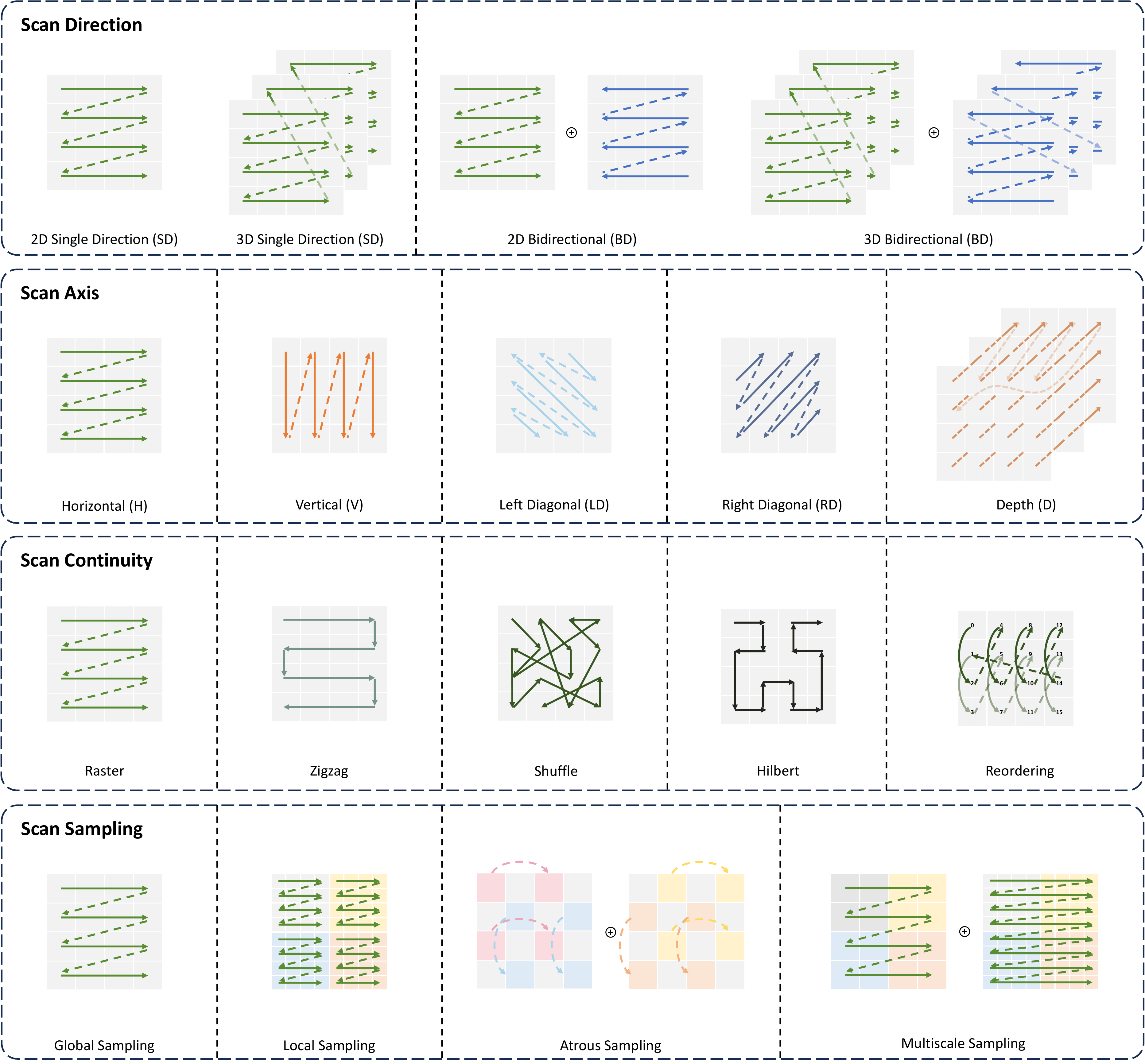}
  \caption{Scanning techniques, categorized into four groups, i.e., scan direction, scan axis, scan continuity, and scan sampling.}
  \Description{}
  \label{fig:scan}
\end{figure}

\subsection{Architectural Advancements}
The typical architecture of sequence models first transforms 2D images into sequences and then applies the sequence transformations. Following this, the architectural advancements of visual Mamba backbone networks focus on tokenization methods, scanning techniques, and architecture designs.

\subsubsection{Tokenization}
The 2D images are first converted into sequences of visual tokens via a stem module, typically comprising a convolution layer followed by a linear projection layer, to be processed. Given the inherent causal properties of SSM operations, adding positional embeddings to visual tokens is optional. 

Some approaches treat these visual tokens as 1D structures, applying 1D SSM transformations and 1D convolution operations, as demonstrated in Vim~\cite{icml24/vim}. Other approaches preserve the 2D structure of visual tokens and adapt operations in Mamba-based blocks to 2D, as seen in VMamba~\cite{neurips24/vmamba}. In addition to visual tokens, other token types can be concatenated or added. Some works insert extra tokens to act as registers~\cite{iclr24/reg} or to perceive discontinuities between rows or columns.

Differently, GlobalMamba~\cite{arxiv24/GlobalMamba} segments images into multiple frequency bands and spatially downsamples them based on their frequency ranges, with lower resolutions for bands in smaller frequency ranges. These bands are then tokenized into causal sequences using a lightweight CNN. Through this, the 2D images are transformed into sequences of causal tokens containing global information.

\subsubsection{Scanning Techniques}
The selective scanning mechanism is the key component of Mamba. However, its original design for 1D causal sequences poses challenges when adapting it to non-causal visual data. Significant research efforts have focused on developing scanning techniques to overcome these challenges. To provide a clear understanding, we categorize these scanning techniques into four main groups: scan direction, scan axis, scan continuity, and scan sampling, forming a decoupled taxonomy. This categorization also reveals different objectives of the scanning techniques. The scan direction addresses the non-causal characteristics of visual sequences; the scan axis deals with the high dimensionality inherent in visual data; the scan continuity considers the spatial continuity of the patches along the scanning path; the scan sampling divides the full image into sub-images to capture spatial information. The illustration of these four groups is shown in Fig.~\ref{fig:scan}.

\paragraph{Scan Direction}
After unfolding the visual data into sequences, different scan directions can be employed to handle the non-causal nature of these sequences. 

Single-directional scanning (\textbf{SD}) represents the original selective scanning technique in Mamba~\cite{arxiv23/mamba}, which processes sequences in a single direction along sequence length. Bi-directional scanning (\textbf{BD}) processes the sequences from the forward and backward directions, thus enhancing the receptive fields reciprocally. 

\paragraph{Scan Axis} Visual data differ from typical sequences by possessing 2D or higher-dimensional spatial information that encapsulates both local and global contexts. Current approaches unfold visual tokens along various axes of the original visual data to thoroughly integrate this spatial information. For instance, the scanning axes for a 2D image typically include the horizontal (\textbf{H}), vertical (\textbf{V}), left diagonal (\textbf{LD}), and right diagonal (\textbf{RD}). For 3D visual data, the axes also involve the depth (\textbf{D}) dimension.
Additionally, Vim-F~\cite{arxiv24/vimf} applies the Fourier Transform to convert the image into the frequency domain (\textbf{F}), enabling a global receptive field during the scanning. 

Beyond the spatial axes, certain approaches apply scanning along the channel (\textbf{C}) dimension to capture additional channel-specific information.

\paragraph{Scan Continuity} Alternative techniques are employed to handle the 2D spatial information in visual data. 

\textbf{Raster} scanning flattens images into sequences following the layout of computer memory. However, this technique can cause spatial discontinuities between adjacent rows or columns, as spatial neighboring tokens in the image may become separated in the sequence. In contrast, \textbf{Zigzag} scanning serializes images following a zigzag path, alternating between left-to-right and right-to-left directions across rows (or columns). This ensures continuous scanning by maintaining spatial continuity among adjacent tokens. Some approaches utilize fractal scanning curves to better capture the structural information within the image. Specifically, \textbf{Hilbert} scanning leverages the Hilbert curve, a space-filling curve that recursively subdivides the image and traverses its regions in a structured manner. This technique effectively preserves the spatial relationships between tokens, significantly enhancing their adjacency within the sequence. \textbf{Reordering} scanning divides the sequence into non-overlapping segments and successively samples instances from each segment in a fixed order, ensuring systematic coverage of the entire sequence. Differently, \textbf{Shuffle} scanning randomizes the token sequence order to enhance positional transformation invariance.

\paragraph{Scan Sampling} In addition to operating on the full image, referred to as \textbf{Global} Sampling, various other sampling techniques are employed for specific purposes.

\textbf{Local} Sampling divides the image into sub-images and performs scanning within each, improving the model's ability to capture local dependencies. \textbf{Atrous} Sampling samples the image into sub-images with skipping, capturing global dependencies while reducing computational complexity. \textbf{Multi-scale} Sampling uses depthwise convolution to generate multi-scale feature maps for scanning. The shorter sequences of low-resolution feature maps help mitigate the long-range forgetting problem while reducing computational cost. \textbf{Quadtree} Sampling uses a quadtree-based image partitioning strategy that estimates the locality score of each token based on its features. Tokens are then adaptively divided into window quadrants at both coarse and fine granularities, enabling the capture of local dependencies across multiple scales. \textbf{Channel} Sampling divides the image channels into groups to improve computational efficiency. 
Notably, within these sub-regions, different combinations of the former three groups of scanning techniques can be applied, allowing for diverse processing approaches without the necessity for uniformity across all sub-regions. 

These four groups of scanning techniques are interoperable and can be synergistically combined to enhance visual data analysis. 

\subsubsection{Architecture}
The core components of the visual Mamba backbone network architecture are various blocks, formed by different combinations of the previously discussed tokenization methods, scanning techniques, and the selective SSM transformation. 

These network architectures can consist of pure Mamba blocks that solely rely on the Mamba architecture for processing visual data. Vim~\cite{icml24/vim} is a pure Mamba network structured as a series of identical Vim blocks, which is a Mamba block integrating a backward SSM path alongside the forward one. VMamba~\cite{neurips24/vmamba} is another pure Mamba network. Its block is similar to the Mamba block but replaces the 1D convolution layer with a 2D depth-wise convolution layer and the selective SSM with 2D-Selective-Scan (SS2D). SS2D unfolds the input patches into sequences along four distinct scanning paths (Cross-Scan), processes each sequence using separate selective SSM transformations in parallel, and then reshapes and combines the resulting sequences to obtain the final output map (Cross-Merge). The PlainMamba~\cite{bmvc24/plainmamba} block is similar to the VMamba block, with two key differences. Firstly, it employs the continuous zigzag scanning technique to maintain spatial adjacency among tokens and prevent discontinuities. Secondly, a direction-aware updating technique is introduced to explicitly incorporate relative 2D position information into the selective scanning process. Differently, Mamba-ND \cite{eccv24/Mamba-ND} treats the 1D Mamba layer as a black box and explores how to unravel and order multi-dimensional data. Its results show that a chain of Mamba layers and simple alternating-directional orderings achieve superior performance. 

\begin{figure}[!t]
  \centering
  \includegraphics[width=0.8\textwidth]{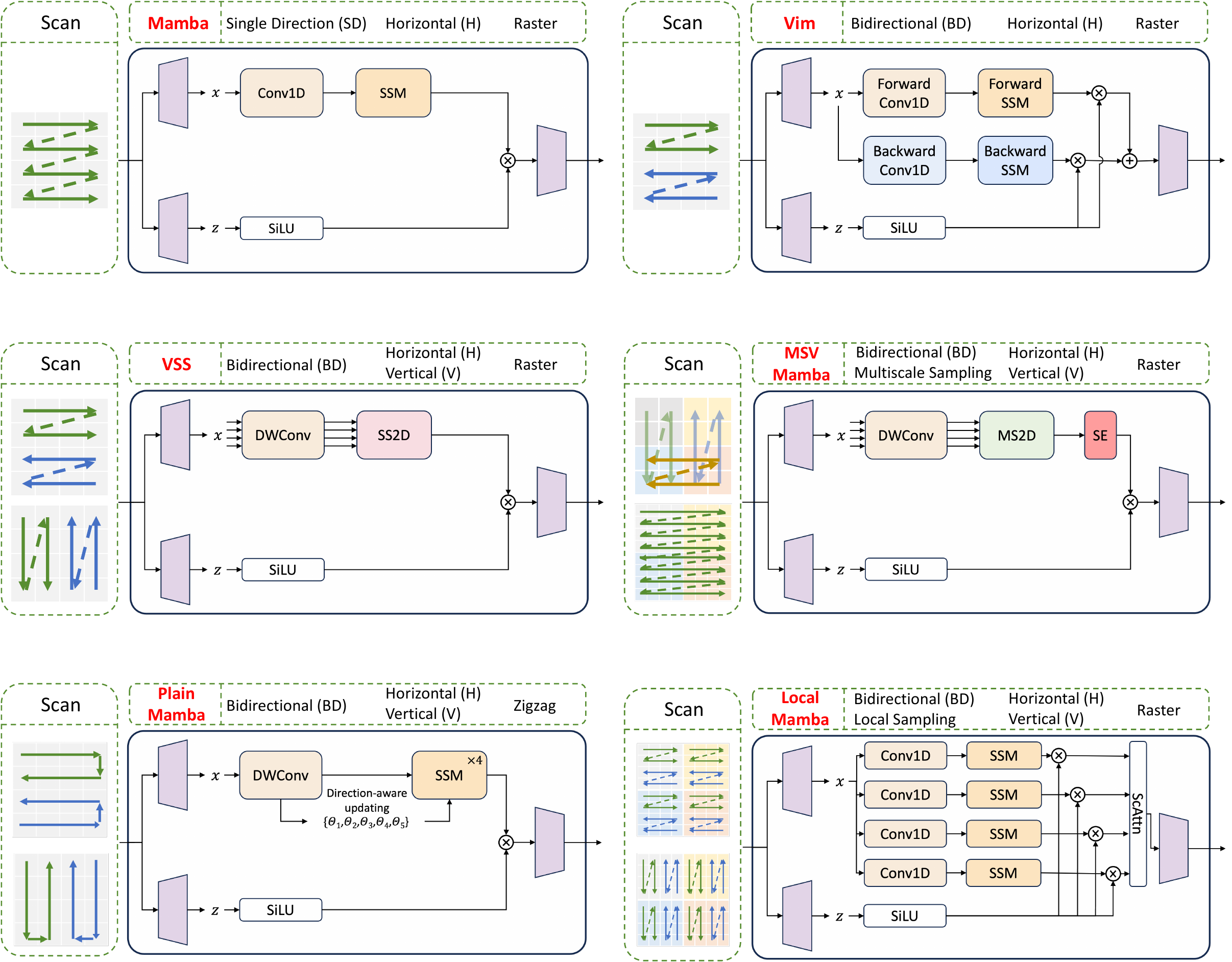}
  \caption{Visual Mamba blocks, including Vision Mamba (Vim)~\cite{icml24/vim}, Visual State Space (VSS)~\cite{neurips24/vmamba}, MSVMamba~\cite{neurips24/MSVMamba}, PlainMamba~\cite{bmvc24/plainmamba}, and LocalMamba~\cite{arxiv24/localmamba} blocks. The original Mamba~\cite{arxiv23/mamba} block
  is presented as a reference for the advancements in the visual Mamba blocks. The scanning techniques and their decoupling results are displayed to the left and above the corresponding blocks, respectively.}
  \Description{}
  \label{fig:backbone_1}
\end{figure}

\begin{table}[!t]
\centering
\caption{Architectural Advancements for Visual Mamba Backbone Networks.}
\resizebox{0.95\textwidth}{!}{
\begin{tabular}{lccccccccc}
\toprule
\multirow{2}{*}{\textbf{Methods}} 
& \multirow{2}{*}{\textbf{Tokenization}} 
& \multicolumn{4}{c}{\textbf{Scan Strategy}}                             
& \multicolumn{2}{c}{\textbf{Architecture}}
& \multirow{2}{*}{\textbf{Venue}}
& \multirow{2}{*}{\textbf{Code}}
\\ \cmidrule(r){3-6} \cmidrule(l){7-8}
                  
&                               
& \textbf{Direction}
& \textbf{Axis}
& \textbf{Continuity}
& \textbf{Sampling} 
& \textbf{Block}
& \textbf{Structure} 
&
&
\\ \midrule
Vim~\cite{icml24/vim} 
& 1D                            
& BD        
& H    
& Raster     
& Global   
& Pure  
& Plain 
& ICML24
& \href{https://github.com/hustvl/Vim}{\color{magenta}\Checkmark}
\\ 
VMamba~\cite{neurips24/vmamba} 
& 2D                            
& BD        
& H/V    
& Raster     
& Global   
& Pure  
& Hierarchical 
& NeurIPS24
& \href{https://github.com/MzeroMiko/VMamba}{\color{magenta}\Checkmark}
\\
Mamba-ND~\cite{eccv24/Mamba-ND} 
& 1D                            
& BD        
& H/V/D    
& Raster     
& Global   
& Pure  
& Plain 
& ECCV24 Oral
& \href{https://github.com/jacklishufan/Mamba-ND}{\color{magenta}\Checkmark}
\\
LocalMamba~\cite{arxiv24/localmamba} 
& 1D 
& BD
& H/V    
& Raster     
& Local   
& Hybrid  
& Plain/Hierarchical 
& 
& \href{https://github.com/hunto/LocalMamba}{\color{magenta}\Checkmark}
\\
EfficientVMamba~\cite{arxiv24/localmamba} 
& 2D 
& BD
& H/V    
& Raster     
& Atrous   
& Hybrid  
& Hierarchical 
& 
& \href{https://github.com/TerryPei/EfficientVMamba}{\color{magenta}\Checkmark}
\\
SiMBA~\cite{arxiv24/simba} 
& 1D 
& SD
& H    
& Raster     
& Global   
& Hybrid  
& Plain 
& 
& \href{https://github.com/badripatro/Simba}{\color{magenta}\Checkmark}
\\
PlainMamba~\cite{bmvc24/plainmamba} 
& 2D 
& BD
& H/V    
& Zigzag     
& Global   
& Pure  
& Plain 
& BMVC24
& \href{https://github.com/ChenhongyiYang/PlainMamba}{\color{magenta}\Checkmark}
\\
MSVMamba~\cite{neurips24/MSVMamba} 
& 2D 
& BD
& H/V    
& Raster     
& Multi-scale   
& Hybrid  
& Hierarchical 
& NeurIPS24
& \href{https://github.com/YuHengsss/MSVMamba}{\color{magenta}\Checkmark}
\\
FractalMamba~\cite{arxiv24/FractalMamba} 
& 2D 
& BD
& H/V    
& Hilbert     
& Global   
& Pure  
& Plain 
& 
& 
\\
Mamba$^{\circledR}$~\cite{arxiv24/Mamba-R} 
& 1D + Register 
& BD
& H    
& Raster     
& Global   
& Pure  
& Plain 
& 
& \href{https://github.com/wangf3014/Mamba-Reg}{\color{magenta}\Checkmark}
\\
Vim-F~\cite{arxiv24/vimf} 
& 1D 
& BD
& H + F    
& Raster     
& Global   
& Pure  
& Plain 
& 
& \href{https://github.com/yws-wxs/Vim-F}{\color{magenta}\Checkmark}
\\
MambaVision~\cite{arxiv24/MambaVision} 
& 1D 
& SD
& H    
& Raster     
& Global   
& Hybrid  
& Hierarchical 
& 
& \href{https://github.com/NVlabs/MambaVision}{\color{magenta}\Checkmark}
\\
GroupMamba~\cite{arxiv24/GroupMamba} 
& 1D 
& BD
& H/V    
& Zigzag     
& Channel   
& Hybrid  
& Hierarchical 
& 
& \href{https://github.com/Amshaker/GroupMamba}{\color{magenta}\Checkmark}
\\
ShuffleMamba~\cite{arxiv24/ShuffleMamba} 
& 1D 
& BD
& H    
& Raster/Shuffle     
& Global   
& Pure  
& Plain 
& 
& \href{https://github.com/huangzizheng01/ShuffleMamba}{\color{magenta}\Checkmark}
\\
StableMamba~\cite{arxiv24/StableMamba} 
& 1D 
& BD
& H    
& Raster     
& Global   
& Hybrid  
& Plain 
& 
& 
\\
QuadMamba~\cite{neurips24/QuadMamba} 
& 2D 
& BD
& H/V    
& Raster     
& Quadtree   
& Pure  
& Hierarchical 
& NeurIPS24
& \href{https://github.com/VISION-SJTU/QuadMamba}{\color{magenta}\Checkmark}
\\
GlobalMamba~\cite{arxiv24/GlobalMamba} 
& F-based Global
& BD
& H/V    
& Raster     
& Global   
& Pure  
& Hierarchical 
& 
& \href{https://github.com/wangck20/GlobalMamba}{\color{magenta}\Checkmark}
\\
\bottomrule
\end{tabular}
}
\label{arc_advance}
\end{table}

The network architectures can also incorporate hybrid Mamba blocks that combine the Mamba architecture with other neural network architectures, such as CNNs and attentions, to capitalize on their complementary advantages. MSVMamba~\cite{neurips24/MSVMamba} incorporates a convolutional feed-forward network in each block to enhance channel-wise information exchange and local feature extraction. LocalMamba~\cite{arxiv24/localmamba} introduces a spatial and channel attention module before patch merging to enhance the integration of directional features and reduce redundancy. 
SiMBA~\cite{arxiv24/simba} uses Mamba for sequence modeling and introduces a new channel modeling technique based on Fourier Transforms.

Fig.~\ref{fig:backbone_1} illustrates a suite of pure and hybrid visual Mamba blocks, including Vision Mamba (Vim)~\cite{icml24/vim}, Visual State Space (VSS)~\cite{neurips24/vmamba}, MSVMamba~\cite{neurips24/MSVMamba}, PlainMamba~\cite{bmvc24/plainmamba}, and LocalMamba~\cite{arxiv24/localmamba} blocks. The Mamba~\cite{arxiv23/mamba} block is also included to facilitate a direct comparison, highlighting the evolutionary design of these blocks in the visual domain. The detailed illustrations of scanning techniques of different blocks and their decoupled results presented in Fig.~\ref{fig:backbone_1} also validate the logic behind our categorization of scanning techniques. 

\begin{wraptable}{r}{0.5\textwidth}
\centering
\caption{Optimization Strategies for Visual Mamba Backbone Networks.}
\resizebox{0.5\textwidth}{!}{
\begin{tabular}{lccc}
\toprule
\textbf{Methods} 
& \textbf{Strategy}                         
& \textbf{Venue}
& \textbf{Code}
\\ \midrule
Mamba-prune~\cite{neurips24/mamba-prune} 
& Token Pruning
& NeurIPS24
& 
\\
VSSD~\cite{arxiv24/VSSD} 
& Visual State Space Duality
& 
& \href{https://github.com/YuHengsss/VSSD}{\color{magenta}\Checkmark}
\\
V2M~\cite{arxiv24/v2m} 
& 2D SSM
& 
& \href{https://github.com/wangck20/V2M}{\color{magenta}\Checkmark}
\\
Spatial-Mamba~\cite{arxiv24/Spatial-Mamba} 
& Structure-aware State Fusion
& 
& \href{https://github.com/EdwardChasel/Spatial-Mamba}{\color{magenta}\Checkmark}
\\
SparX-Mamba~\cite{arxiv24/SparX} 
& Cross-layer Connection
& 
& \href{https://github.com/LMMMEng/SparX}{\color{magenta}\Checkmark}
\\
HRVMamba~\cite{arxiv24/HRVMamba} 
& Multi-resolution Parallel Design
& 
& \href{https://github.com/zhanghao5201/HRVMamba}{\color{magenta}\Checkmark}
\\
ARM~\cite{arxiv24/ARM} 
& Autoregressive Pre-training
& 
& \href{https://github.com/OliverRensu/ARM}{\color{magenta}\Checkmark}
\\
MAP~\cite{arxiv24/MAP} 
& Masked Autoregressive Pre-training
& 
& 
\\
\bottomrule
\end{tabular}
}
\label{opt_strategies}
\end{wraptable}

Built on different types of blocks, existing architectures also differ in their use of plain or hierarchical structures. Plain structures typically stack identical blocks, while hierarchical structures introduce down-sampling layers between block stacks to generate hierarchical representations. 

Details of tokenization methods, scanning techniques, and architecture designs of visual Mamba backbone networks are outlined in Table.~\ref{arc_advance}.

\subsection{Optimization Strategies}
In addition to architectural advancements, researchers also explore additional optimization strategies to enhance the efficiency and performance of visual Mamba backbone networks. A summary of these strategies is presented in Table.~\ref{opt_strategies}. Mamba-prune~\cite{neurips24/mamba-prune} introduces token pruning into Mamba-based networks for accelerating the computation. VSSD~\cite{arxiv24/VSSD} adapts SSD to a non-causal mode to better handle the non-causality of visual data, while V2M~\cite{arxiv24/v2m} and Spatial-Mamba~\cite{arxiv24/Spatial-Mamba} extend the SSM to 2D forms. SparX-Mamba~\cite{arxiv24/SparX} and HRVMamba~\cite{arxiv24/HRVMamba} introduce additional designs for cross-layer feature aggregation and dense prediction, respectively. ARM~\cite{arxiv24/ARM} and MAP~\cite{arxiv24/MAP} focus on improving the model performance via pre-training.

\begin{table}[!t]
\centering
\caption{Comparison of different backbones on ImageNet-1K \cite{cvpr09/imagenet_1k} classification at the resolution of $224^2$ (Part I). `Conv', `Attn', and `Mamba' denote architectures based on CNN, Transformer, and Mamba, respectively. 
$^{\star}$ indicates that the backbone has undergone additional training using the Token Labeling objective \cite{aaai22/scaled_relu}. $^{\ast}$ indicates results are obtained with MESA \cite{neurips22/MESA}. $^{\dagger}$ indicates results reproduced by DeiT \cite{icml21/deit}. $^{\ddagger}$ indicates hierarchical structures while others are plain structures. $^l$ represents local sampling. The \underline{\textbf{best}} results are marked with bold and underline, the \textbf{second-best} with bold only, and the \underline{third-best} with underline only.
}
\label{tab:backbone_comparison}
\begin{tabularx}{\textwidth}{X X}
\centering
\resizebox{0.48\textwidth}{!}{
\begin{tabular}	{l c c c c}
\toprule
\textbf{Backbone} & \textbf{Type} & \textbf{Params} & \textbf{FLOPs} & \textbf{Top-1 ACC} \\
\midrule
\textbf{DeiT-Ti} \cite{icml21/deit} & Attn & 5.0 M & 1.3 G & 72.2 \\ 
\textbf{EffVMamba-T} \cite{arxiv24/effvmamba} & Mamba & 6.0 M & 0.8 G & 76.5 \\
\textbf{QuadMamba-Li} \cite{neurips24/QuadMamba} & Mamba & 5.4 M & 0.8 G & 74.2 \\
\textbf{MSVMamba-N} \cite{neurips24/MSVMamba} & Mamba & 7.0 M & 0.9 G & \textbf{77.3} \\ 
\midrule
\textbf{Vim-Ti} \cite{icml24/vim} & Mamba & 7.0 M & 1.5 G & 76.1 \\
\textbf{Mamba$^{\circledR}$-T} \cite{arxiv24/Mamba-R} & Mamba & 9.0 M & 5.1 G & 77.4 \\
\textbf{PlainMamba-L1} \cite{bmvc24/plainmamba} & Mamba & 7.0 M & 3.0 G & 77.9 \\ 
\textbf{LocalVim-T} \cite{arxiv24/localmamba} & Mamba & 8.0 M & 1.5 G & 76.2 \\
\textbf{EffVMamba-S} \cite{arxiv24/effvmamba} & Mamba & 11.0 M & 1.3 G & 78.7 \\
\textbf{V2M-T} \cite{arxiv24/v2m} & Mamba & 7 M & 1.9 G & 76.2 \\
\textbf{V2M-T$^{l}$} \cite{arxiv24/v2m} & Mamba & 8 M & 1.8 G & 76.4 \\
\textbf{GlobalMamba-M} \cite{arxiv24/GlobalMamba} & Mamba & 7 M & 1.7 G & 76.4 \\
\textbf{QuadMamba-T} \cite{neurips24/QuadMamba} & Mamba & 10 M & 2.0 G & 78.2 \\
\textbf{MSVMamba-M} \cite{neurips24/MSVMamba} & Mamba & 12.0 M & 1.5 G & \underline{79.8} \\
\textbf{SiMBA-S} \cite{arxiv24/simba} & Mamba & 15.3 M & 2.4 G & \textbf{81.7} \\
\textbf{VSSD-M} \cite{arxiv24/VSSD} & Mamba & 14 M & 2.3 G & \underline{\textbf{82.5}} \\
\midrule
\textbf{RegNetY-4G}$^{\dagger}$ \cite{cvpr20/regnet} & Conv & 20.6 M & 4.0 G & 80.0 \\
\textbf{ConvNeXt-T} \cite{cvpr22/convnext} & Conv & 29.0 M & 4.5 G & 82.1 \\
\textbf{MambaOut-T} \cite{arxiv24/mambaout} & Conv & 27.0 M & 4.5 G & 82.7 \\
\textbf{DeiT-S} \cite{icml21/deit} & Attn & 22.0 M & 4.6 G & 79.8 \\ 
\textbf{Swin-T} \cite{iccv21/swinvit} & Attn & 29.0 M & 4.5 G & 81.3 \\
\textbf{Wave-ViT-S}$^{\star}$ \cite{eccv22/wavevit} & Attn & 22.7 M & 4.7 G & \underline{83.9} \\ 
\textbf{SVT-H-S}$^{\star}$ \cite{neurips23/svt} & Attn & 22.0 M & 3.9 G & \underline{\textbf{84.2}} \\ 
\textbf{VOLO-D1}$^{\star}$ \cite{pami23/volo} & Attn & 27.0 M & 6.8 G & \underline{\textbf{84.2}} \\ 
\hdashline
\textbf{Vim-S} \cite{icml24/vim} & Mamba & 26.0 M & 5.1 G & 80.5 \\
\textbf{Mamba$^{\circledR}$-S} \cite{arxiv24/Mamba-R} & Mamba & 28.0 M & 9.9 G & 81.1 \\
\textbf{PlainMamba-L2} \cite{bmvc24/plainmamba} & Mamba & 25.0 M & 8.1 G & 81.6 \\
\textbf{Mamba-ND-S} \cite{eccv24/Mamba-ND} & Mamba & 24.0 M & - & 81.7 \\ 
\textbf{VMamba-T} \cite{neurips24/vmamba} & Mamba & 31.0 M & 4.9 G & 82.5 \\
\textbf{FractalMamba-T} \cite{arxiv24/FractalMamba} & Mamba & 31.0 M & 4.9 G & 82.7 \\ 
\textbf{LocalVim-S} \cite{arxiv24/localmamba} & Mamba & 28.0 M & 4.8 G & 81.2 \\
\textbf{LocalVMamba-T} \cite{arxiv24/localmamba} & Mamba & 26.0 M & 5.7 G & 82.7 \\
\textbf{EffVMamba-B} \cite{arxiv24/effvmamba} & Mamba & 33.0 M & 4.0 G & 81.8 \\
\textbf{V2M-S} \cite{arxiv24/v2m} & Mamba & 26 M & 5.9 G & 80.5 \\
\textbf{V2M-S$^{l}$} \cite{arxiv24/v2m} & Mamba & 28 M & 5.4 G & 81.3 \\
\textbf{V2M-S$^{\ddagger}$} \cite{arxiv24/v2m} & Mamba & 30 M & 5.4 G & 82.9 \\
\textbf{GlobalMamba-T} \cite{arxiv24/GlobalMamba} & Mamba & 26 M & 5.7 G & 80.8 \\
\textbf{GlobalMamba-T$^{\ddagger}$} \cite{arxiv24/GlobalMamba} & Mamba & 30 M & 5.3 G & 82.8 \\
\textbf{QuadMamba-S} \cite{neurips24/QuadMamba} & Mamba & 31 M & 5.5 G & 82.4 \\
\textbf{MSVMamba-T} \cite{neurips24/MSVMamba} & Mamba & 33.0 M & 4.6 G & 82.8 \\
\textbf{SiMBA-B} \cite{arxiv24/simba} & Mamba & 22.8 M & 4.2 G & 83.5 \\
\textbf{MambaVision-T} \cite{arxiv24/MambaVision} & Mamba & 31.8 M & 4.4 G & 82.3 \\
\textbf{MambaVision-T2} \cite{arxiv24/MambaVision} & Mamba & 35.1 M & 5.1 G & 82.7 \\ 
\textbf{GroupMamba-T} \cite{arxiv24/GroupMamba} & Mamba & 23 M & 4.6 G & 83.3 \\
\textbf{Spatial-Mamba-T} \cite{arxiv24/Spatial-Mamba} & Mamba & 27 M & 4.5 G & 83.5 \\
\textbf{VSSD-T} \cite{arxiv24/VSSD} & Mamba & 24 M & 4.5 G & 83.7 \\
\textbf{VSSD-T}$^{\ast}$ \cite{arxiv24/VSSD} & Mamba & 24 M & 4.5 G & \textbf{84.1} \\
\bottomrule
\end{tabular}
}
&
\centering
\resizebox{0.48\textwidth}{!}{
\begin{tabular}	{l c c c c}
\toprule
\textbf{Backbone} & \textbf{Type} & \textbf{Params} & \textbf{FLOPs} & \textbf{Top-1 ACC} \\
\midrule
\textbf{RegNetY-8G}$^{\dagger}$ \cite{cvpr20/regnet} & Conv & 39.2 M & 8.0 G & 81.7 \\
\textbf{ConvNeXt-S} \cite{cvpr22/convnext} & Conv & 50.0 M & 8.7 G & 83.1 \\
\textbf{MambaOut-S} \cite{arxiv24/mambaout} & Conv & 48.0 M & 9.0 G & 84.1 \\
\textbf{Swin-S} \cite{iccv21/swinvit} & Attn & 50.0 M & 8.7 G & 83.0 \\
\textbf{Wave-ViT-B}$^{\star}$ \cite{eccv22/wavevit} & Attn & 33.5 M & 7.2 G & \textbf{84.8} \\ 
\textbf{SVT-H-B}$^{\star}$ \cite{neurips23/svt} & Attn & 32.8 M & 6.3 G & \underline{\textbf{85.2}} \\ 
\textbf{VOLO-D2}$^{\star}$ \cite{pami23/volo} & Attn & 59.0 M & 14.1 G & \underline{\textbf{85.2}} \\ 
\hdashline
\textbf{PlainMamba-L3} \cite{bmvc24/plainmamba} & Mamba & 50.0 M & 14.4 G & 82.3 \\
\textbf{VMamba-S} \cite{neurips24/vmamba} & Mamba & 50.0 M & 8.7 G & 83.6 \\
\textbf{LocalVMamba-S} \cite{arxiv24/localmamba} & Mamba & 50.0 M & 11.4 G & 83.7 \\
\textbf{V2M-B$^{\ddagger}$} \cite{arxiv24/v2m} & Mamba & 50 M & 9.6 G & 83.8 \\
\textbf{GlobalMamba-S$^{\ddagger}$} \cite{arxiv24/GlobalMamba} & Mamba & 50 M & 9.5 G & 83.9 \\
\textbf{QuadMamba-B} \cite{neurips24/QuadMamba} & Mamba & 50 M & 9.3 G & 83.8 \\
\textbf{SiMBA-L} \cite{arxiv24/simba} & Mamba & 36.6 M & 7.6 G & 84.4 \\ 
\textbf{MambaVision-S} \cite{arxiv24/MambaVision} & Mamba & 50.1 M & 7.5 G & 83.3 \\ 
\textbf{GroupMamba-S} \cite{arxiv24/GroupMamba} & Mamba & 34 M & 7.0 G & 83.9 \\
\textbf{Spatial-Mamba-S} \cite{arxiv24/Spatial-Mamba} & Mamba & 43 M & 7.1 G & \underline{84.6} \\
\textbf{VSSD-S} \cite{arxiv24/VSSD} & Mamba & 40 M & 7.4 G & 84.1 \\
\textbf{VSSD-S}$^{\ast}$ \cite{arxiv24/VSSD} & Mamba & 40 M & 7.4 G & 84.5 \\
\midrule
\textbf{RegNetY-16G}$^{\dagger}$ \cite{cvpr20/regnet} & Conv & 83.6 M & 16.0 G & 82.9 \\
\textbf{ConvNeXt-B} \cite{cvpr22/convnext} & Conv & 89.0 M & 15.4 G & 83.8 \\
\textbf{MambaOut-B} \cite{arxiv24/mambaout} & Conv & 85.0 M & 15.8 G & 84.2 \\
\textbf{DeiT-B} \cite{icml21/deit} & Attn & 86.0 M & 17.5 G & 81.8 \\ 
\textbf{Swin-B} \cite{iccv21/swinvit} & Attn & 88.0 M & 15.4 G & 83.5 \\
\textbf{Wave-ViT-L}$^{\star}$ \cite{eccv22/wavevit} & Attn & 57.5 M & 14.8 G & \textbf{85.5} \\ 
\textbf{SVT-H-L}$^{\star}$ \cite{neurips23/svt} & Attn & 54.0 M & 12.7 G & \underline{\textbf{85.7}} \\ 
\textbf{VOLO-D3}$^{\star}$ \cite{pami23/volo} & Attn & 86.0 M & 20.6 G & 85.4 \\ 
\hdashline
\textbf{Mamba$^{\circledR}$-B} \cite{arxiv24/Mamba-R} & Mamba & 99.0 M & 20.3 G & 82.9 \\
\textbf{Mamba-ND-B} \cite{eccv24/Mamba-ND} & Mamba & 92.0 M & - & 83.0 \\ 
\textbf{VMamba-B} \cite{neurips24/vmamba} & Mamba & 89.0 M & 15.4 G & 83.9 \\ 
\textbf{ARM-B} \cite{arxiv24/ARM} & Mamba & 85 M & - & 83.2 \\
\textbf{GlobalMamba-B$^{\ddagger}$} \cite{arxiv24/GlobalMamba} & Mamba & 89 M & 17.0 G & 84.1 \\
\textbf{MambaVision-B} \cite{arxiv24/MambaVision} & Mamba & 97.7 M & 15.0 G & 84.2 \\ 
\textbf{GroupMamba-B} \cite{arxiv24/GroupMamba} & Mamba & 57 M & 14.0 G & 84.5 \\
\textbf{Spatial-Mamba-B} \cite{arxiv24/Spatial-Mamba} & Mamba & 96 M & 15.8 G & 85.3 \\
\textbf{VSSD-B} \cite{arxiv24/VSSD} & Mamba & 89 M & 16.1 G & 84.7 \\
\textbf{VSSD-B}$^{\ast}$ \cite{arxiv24/VSSD} & Mamba & 89 M & 16.1 G & \underline{85.4} \\
\midrule
\textbf{VOLO-D4}$^{\star}$ \cite{pami23/volo} & Attn & 193.0 M & 43.8 G & \textbf{85.7} \\ 
\textbf{ARM-L} \cite{arxiv24/ARM} & Mamba & 297 M & - & 84.5 \\
\textbf{MambaVision-L} \cite{arxiv24/MambaVision} & Mamba & 227.9 M & 34.9 G & 85.0 \\ 
\textbf{MambaVision-L2} \cite{arxiv24/MambaVision} & Mamba & 241.5 M & 37.5 G & 85.3 \\
\midrule
\textbf{VOLO-D5}$^{\star}$ \cite{pami23/volo} & Attn & 296.0 M & 69.0 G & \textbf{86.1} \\ 
\textbf{Mamba$^{\circledR}$-L} \cite{arxiv24/Mamba-R} & Mamba & 341.0 M & 55.5 G & 83.2 \\
\textbf{ARM-H} \cite{arxiv24/ARM} & Mamba & 662 M & - & 85.0 \\
\bottomrule
\end{tabular}
}
\end{tabularx}
\end{table}

\begin{figure}[!t]
  \centering
  \includegraphics[width=0.6\textwidth]{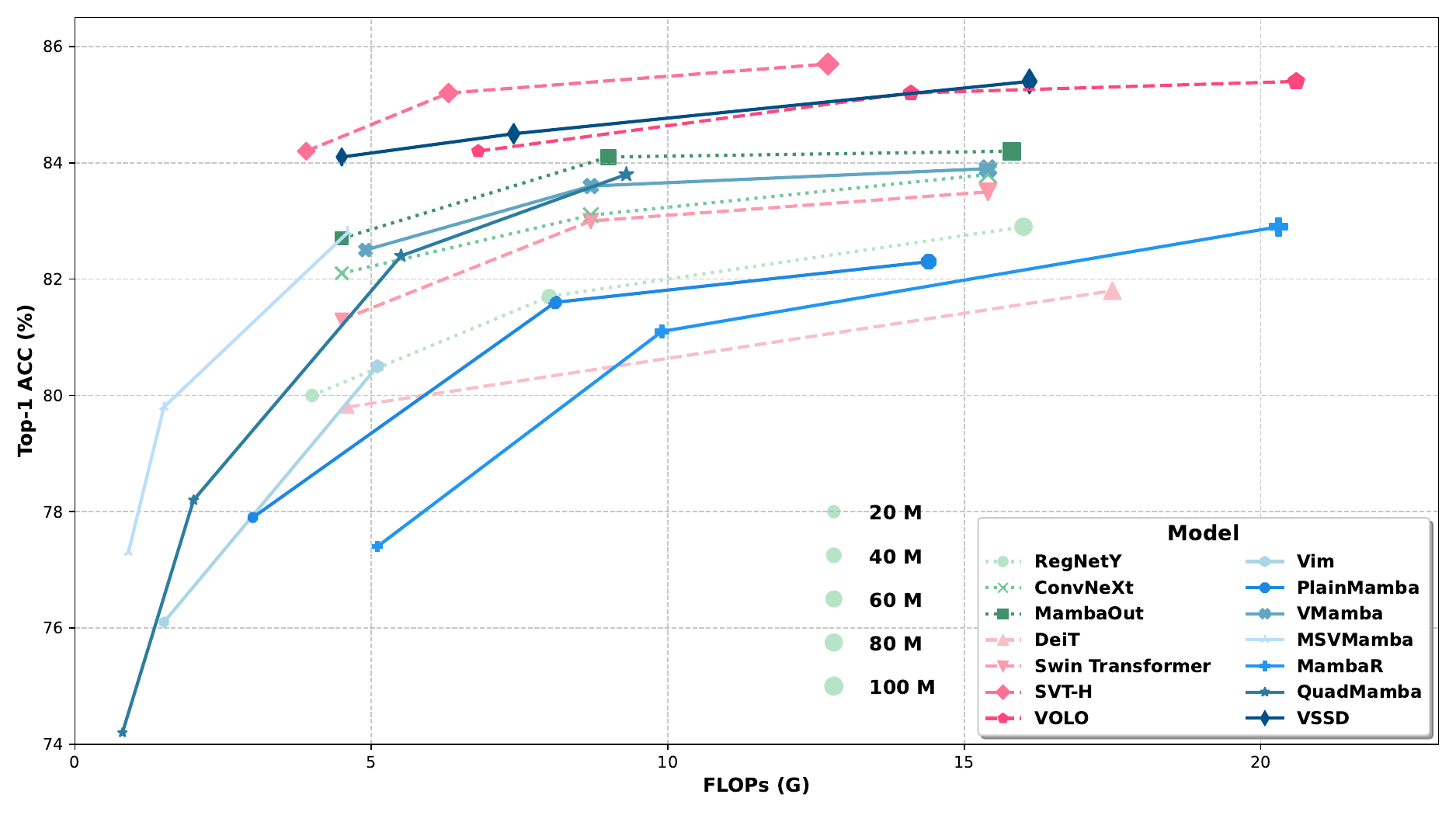}
  \caption{Comparative analysis of the performance and computational complexity across various visual backbone architectures, encompassing Convolution-based methods~\cite{cvpr20/regnet,cvpr22/convnext, arxiv24/mambaout}, Transformer-based methods~\cite{icml21/deit,iccv21/swinvit,eccv22/wavevit,neurips23/svt,pami23/volo}, and Mamba-based methods~\cite{icml24/vim,bmvc24/plainmamba,neurips24/vmamba,neurips24/MSVMamba,arxiv24/Mamba-R,neurips24/QuadMamba,arxiv24/VSSD}. 
  The symbol size is proportional to the parameter count of respective model, providing a visual indicator of model scale and complexity. Note that all data are accessed from released academic papers to ensure fairness and credibility.}
  \Description{}
  \label{fig:cls_compare}
\end{figure}

\subsection{Experimental Results}
In this section, we compare the experimental results of different visual Mamba backbone networks with established architectures based on CNNs or Transformers on standard computer vision benchmarks.

\subsubsection{Image Classification}
Image classification performance is compared on ImageNet-1K \cite{cvpr09/imagenet_1k} and the Top-1 accuracy results from the respective papers are listed in Table.~\ref{tab:backbone_comparison}. 
To clearly compare different networks, we also plot their performance and computational complexity in Fig.~\ref{fig:cls_compare}. 
As seen, the performance of all the models improves as model sizes increase, illustrating the trend of enhancing performance through model scaling. It is evident that with similar computational complexity, all visual Mamba networks outperform the CNN-based RegNetY \cite{cvpr20/regnet} and the Transformer-based DeiT \cite{icml21/deit}. Furthermore, many visual Mamba networks surpass the CNN-based MambaOut \cite{arxiv24/mambaout} and Transformer-based Swin Transformer \cite{iccv21/swinvit}. These results demonstrate the superior performance of visual Mamba networks compared to advanced CNNs and common Transformers. However, Transformer-based networks are highly competitive, particularly SVT \cite{neurips23/svt}, Wave-ViT \cite{eccv22/wavevit}, and VOLO \cite{pami23/volo}, which consistently demonstrate superior performance over other networks. Notably, VOLO \cite{pami23/volo} also exhibits excellent scalability. 
Currently, despite their efficiency, most Mamba-based networks are limited to small-scale implementations with FLOPs under 21G. While Mamba$^{\circledR}$~\cite{arxiv24/Mamba-R} and ARM~\cite{arxiv24/ARM} have demonstrated scalability to larger models, there remains a need for performance enhancements. 
To sum up, visual Mamba networks exhibit promising performance but still underperform advanced Transformer-based networks. Further exploration is also needed to scale visual Mamba networks to larger configurations.

\subsubsection{Object Detection}
Object detection and instance segmentation performance are compared on the MS COCO \cite{eccv14/coco} via Mask R-CNN \cite{iccv17/mask_rcnn} and the results from the respective papers are listed in Table.~\ref{tab:mask_rcnn_coco}. 
The object detection and instance segmentation performance of all the models follows the scaling law as well. It can be observed that with comparable computational complexity, visual Mamba networks surpass the CNN-based ConvNeXt \cite{cvpr22/convnext}, except for EfficientVMamba \cite{arxiv24/effvmamba}, which is designed for lightweight purposes. Most visual Mamba networks demonstrate superior performance compared to all CNN-based architectures and a majority of Transformer-based models. Nonetheless, in certain scenarios, their performance is slightly below that of SG-Former \cite{iccv23/sgformer}. 
These results demonstrate that Mamba's ability to capture long-range dependencies and utilize dynamic weights is advantageous for dense prediction tasks. 
Currently, Mamba-based models are not specifically tailored for these tasks, indicating potential for future research to explore and enhance their performance.

\subsubsection{Semantic Segmentation}
Semantic segmentation performance is compared on ADE20K \cite{ijcv19/ade20k} utilizing UperNet \cite{eccv18/upernet}  and the results from the respective papers are listed in Table.~\ref{tab:upernet_ade20k_seg}. 
The semantic segmentation performance of different networks is similar to their object detection performance. Many visual Mamba networks outperform all CNN-based networks and most Transformer-based networks. However, their performance lags behind that of SG-Former \cite{iccv23/sgformer}. These results further demonstrate Mamba's efficacy in dense prediction tasks. Further exploration of Mamba-based models specifically designed for these tasks is warranted.


\begin{table}[!t]
\centering
\caption{Results of object detection and instance segmentation on MS COCO \cite{eccv14/coco} $mini$-$val$ using Mask R-CNN \cite{iccv17/mask_rcnn} 1$\times$ schedule. `Conv', `Attn', and `Mamba' denote architectures based on CNN, Transformer, and Mamba, respectively. FLOPs are computed using an input size 1280$\times$800. $^{\dagger}$ indicates FLOPs computed using an input size 1333$\times$800. $^{\ddagger}$ indicates hierarchical structures while others are plain structures. The \underline{\textbf{best}} results are marked with bold and underline, the \textbf{second-best} with bold only, and the \underline{third-best} with underline only.}
\label{tab:mask_rcnn_coco}
\resizebox{0.65\textwidth}{!}{
\begin{tabular}{lc|cc|ccc|cccc}
\toprule
\textbf{Backbone} & \textbf{Type} & \textbf{Params} & \textbf{FLOPs} & $\mathbf{AP^{b}}$ & $\mathbf{AP^{b}_{50}}$ & $\mathbf{AP^{b}_{75}}$ & $\mathbf{AP^{m}}$ & $\mathbf{AP^{m}_{50}}$ & $\mathbf{AP^{m}_{75}}$ \\
\midrule
\textbf{ConvNeXt-T} \cite{cvpr22/convnext} & Conv & 48.0 M & 262.0 G & 44.2 & 66.6 & 48.3 & 40.1 & 63.3 & 42.8 \\
\textbf{MambaOut-T} \cite{arxiv24/mambaout} & Conv & 43.0 M & 262.0 G & 45.1 & 67.3 & 49.6 & 41.0 & 64.1 & 44.1 \\
\textbf{Swin-T} \cite{iccv21/swinvit} & Attn & 48.0 M & 264.0 G & 42.2 & - & - & 39.1 & - & - \\
\textbf{PVTv2-B2} \cite{cvm22/pvtv2} & Attn & 45.0 M & - & 45.3 & 67.1 & 49.6 & 41.2 & 64.2 & 44.4 \\ 
\textbf{ViL-S}$^{\dagger}$ \cite{iccv21/vil} & Attn & 45.0 M & 218.3 G & 44.9 & 67.1 & 49.3 & 41.0 & 64.2 & 44.1 \\
\textbf{SG-Former-S} \cite{iccv23/sgformer} & Attn & 41.0 M & - & \underline{47.4} & 69.0 & 52.0 & \underline{42.6} & 65.9 & 46.0 \\
\hdashline
\textbf{PlainMamba-L1} \cite{bmvc24/plainmamba} & Mamba & 31.0 M & 388.0 G & 44.1 & 64.8 & 47.9 & 39.1 & 61.6 & 41.9 \\
\textbf{VMamba-T} \cite{neurips24/vmamba} & Mamba & 50.0 M & 270.0 G & \underline{47.4} & 69.5 & 52.0 & \textbf{42.7} & 66.3 & 46.0 \\
\textbf{FractalMamba-T} \cite{arxiv24/FractalMamba} & Mamba & 50.0 M & 270.0 G & \underline{\textbf{47.8}} & 70.0 & 52.4 & \underline{\textbf{42.9}} & 66.6 & 46.3 \\ 
\textbf{EffVMamba-S} \cite{arxiv24/effvmamba} & Mamba & 31.0 M & 197.0 G & 39.3 & 61.8 & 42.6 & 36.7 & 58.9 & 39.2 \\
\textbf{V2M-S$^{\ddagger}$} \cite{arxiv24/v2m} & Mamba & 50 M & - & \textbf{47.6} & 69.4 & 52.2 & \underline{\textbf{42.9}} & 66.5 & 46.3 \\
\textbf{GlobalMamba-T$^{\ddagger}$} \cite{arxiv24/GlobalMamba} & Mamba & 50 M & - & \textbf{47.6} & 69.4 & 52.5 & \underline{\textbf{42.9}} & 66.5 & 46.0 \\
\textbf{QuadMamba-T} \cite{neurips24/QuadMamba} & Mamba & 30 M & 213 G & 42.3 & 64.6 & 46.2 & 38.8 & 61.6 & 41.4 \\ 
\textbf{MSVMamba-M} \cite{neurips24/MSVMamba} & Mamba & 32.0 M & 201.0 G & 43.8 & 65.8 & 47.7 & 39.9 & 62.9 & 42.9 \\ 
\textbf{LocalVMamba-T} \cite{arxiv24/localmamba} & Mamba & 45.0 M & 291.0 G & 46.7 & 68.7 & 50.8 & 42.2 & 65.7 & 45.5 \\
\textbf{GroupMamba} \cite{arxiv24/GroupMamba} & Mamba & 40 M & 279 G & \textbf{47.6} & 69.8 & 52.1 & \underline{\textbf{42.9}} & 66.5 & 46.3 \\
\textbf{Spatial-Mamba-T} \cite{arxiv24/Spatial-Mamba} & Mamba & 46 M & 261 G & \textbf{47.6} & 69.6 & 52.3 & \underline{\textbf{42.9}} & 66.5 & 46.2 \\
\textbf{VSSD-M} \cite{arxiv24/VSSD} & Mamba & 33 M & 220 G & 45.4 & 67.5 & 49.8 & 41.3 & 64.5 & 44.6 \\
\textbf{VSSD-T} \cite{arxiv24/VSSD} & Mamba & 44 M & 265 G & 46.9 & 69.4 & 51.4 & \underline{42.6} & 66.4 & 45.9 \\
\midrule
\textbf{ConvNeXt-S} \cite{cvpr22/convnext} & Conv & 70.0 M & 348.0 G & 45.4 & 67.9 & 50.0 & 41.8 & 65.2 & 45.1 \\
\textbf{MambaOut-S} \cite{arxiv24/mambaout} & Conv & 65.0 M & 354.0 G & 47.4 & 69.1 & 52.4 & 42.7 & 66.1 & 46.2 \\
\textbf{ViT-Adpt-S} \cite{iclr23/vit_adpt} & Attn & 47.8 M & - & 44.7 & 65.8 & 48.3 & 39.9 & 62.5 & 42.8 \\
\textbf{Swin-S} \cite{iccv21/swinvit} & Attn & 69.0 M & 354.0 G & 44.8 & - & - & 40.9 & - & - \\
\textbf{PVTv2-B3} \cite{cvm22/pvtv2} & Attn & 64.9 M & - & 47.0 & 68.1 & 51.7 & 42.5 & 65.7 & 45.7 \\ 
\textbf{ViL-M}$^{\dagger}$ \cite{iccv21/vil} & Attn & 60.1 M & 293.8 G & 47.6 & 69.8 & 52.1 & 43.0 & 66.9 & 46.6 \\
\textbf{SG-Former-M} \cite{iccv23/sgformer} & Attn & 51.0 M & - & 48.2 & 70.3 & 53.1 & 43.6 & 66.9 & 47.0 \\
\hdashline
\textbf{PlainMamba-L2} \cite{bmvc24/plainmamba} & Mamba & 53.0 M & 542.0 G & 46.0 & 66.9 & 50.1 & 40.6 & 63.8 & 43.6 \\
\textbf{VMamba-S} \cite{neurips24/vmamba} & Mamba & 64.0 M & 357.0 G & 48.7 & 70.0 & 53.4 & 43.7 & 67.3 & 47.0 \\
\textbf{EffVMamba-B} \cite{arxiv24/effvmamba} & Mamba & 53.0 M & 252.0 G & 43.7 & 66.2 & 47.9 & 40.2 & 63.3 & 42.9 \\
\textbf{V2M-B$^{\ddagger}$} \cite{arxiv24/v2m} & Mamba & 70 M & - & \underline{48.9} & 70.2 & 53.6 & \underline{43.8} & 67.5 & 47.1 \\
\textbf{GlobalMamba-S$^{\ddagger}$} \cite{arxiv24/GlobalMamba} & Mamba & 70 M & - & \textbf{49.0} & 70.5 & 53.5 & \textbf{43.9} & 67.5 & 47.0 \\
\textbf{QuadMamba-S} \cite{neurips24/QuadMamba} & Mamba & 55 M & 301 G & 46.7 & 69.0 & 51.3 & 42.4 & 65.9 & 45.6 \\ 
\textbf{MSVMamba-T} \cite{neurips24/MSVMamba} & Mamba & 53.0 M & 252.0 G & 46.9 & 68.8 & 51.4 & 42.2 & 65.6 & 45.4 \\ 
\textbf{LocalVMamba-S} \cite{arxiv24/localmamba} & Mamba & 69.0 M & 414.0 G & 48.4 & 69.9 & 52.7 & 43.2 & 66.7 & 46.5 \\
\textbf{SiMBA-S} \cite{arxiv24/simba} & Mamba & 60.0 M & 382.0 G & 46.9 & 68.6 & 51.7 & 42.6 & 65.9 & 45.8 \\
\textbf{Spatial-Mamba-S} \cite{arxiv24/Spatial-Mamba} & Mamba & 63 M & 315 G & \underline{\textbf{49.2}} & 70.8 & 54.2 & \underline{\textbf{44.0}} & 67.9 & 47.5 \\
\textbf{VSSD-S} \cite{arxiv24/VSSD} & Mamba & 59 M & 325 G & 48.4 & 70.1 & 53.1 & 43.5 & 67.2 & 47.1 \\
\midrule
\textbf{MambaOut-B} \cite{arxiv24/mambaout} & Conv & 100.0 M & 495.0 G & 47.4 & 69.3 & 52.2 & 43.0 & 66.4 & 46.3 \\
\textbf{ViT-Adpt-B} \cite{iclr23/vit_adpt} & Attn & 120.2 M & - & 47.0 & 68.2 & 51.4 & 41.8 & 65.1 & 44.9 \\
\textbf{PVTv2-B4} \cite{cvm22/pvtv2} & Attn & 82.2 M & - & 47.5 & 68.7 & 52.0 & 42.7 & 66.1 & 46.1 \\ 
\textbf{PVTv2-B5} \cite{cvm22/pvtv2} & Attn & 101.6 M & - & 47.4 & 68.6 & 51.9 & 42.5 & 65.7 & 46.0 \\ 
\textbf{Swin-B} \cite{iccv21/swinvit} & Attn & 107.0 M & 496.0 G & 46.9 & - & - & 42.3 & - & - \\ 
\textbf{ViL-B}$^{\dagger}$ \cite{iccv21/vil} & Attn & 76.1 M & 384.4 G & 48.6 & 70.5 & 53.4 & 43.6 & 67.6 & 47.1 \\
\textbf{SG-Former-B} \cite{iccv23/sgformer} & Attn & 95.0 M & - & \underline{49.2} & 70.6 & 54.3 & \underline{\textbf{68.2}} & 68.1 & 47.7 \\
\hdashline
\textbf{PlainMamba-L3} \cite{bmvc24/plainmamba} & Mamba & 79.0 M & 696.0 G & 46.8 & 68.0 & 51.1 & 41.2 & 64.7 & 43.9 \\ 
\textbf{VMamba-B} \cite{neurips24/vmamba} & Mamba & 108.0 M & 485.0 G & \underline{49.2} & 70.9 & 53.9 & 43.9 & 67.7 & 47.6 \\
\textbf{GlobalMamba-B$^{\ddagger}$} \cite{arxiv24/GlobalMamba} & Mamba & 108 M & - & \textbf{49.3} & 71.4 & 54.2 & \underline{44.2} & 68.4 & 47.7 \\
\textbf{Spatial-Mamba-B} \cite{arxiv24/Spatial-Mamba} & Mamba & 115 M & 494 G & \underline{\textbf{50.4}} & 71.8 & 55.3 & \textbf{45.1} & 69.1 & 49.1 \\
\bottomrule
\end{tabular}}
\end{table}

\clearpage

\begin{table}[!t]
\centering
\caption{Results of semantic segmentation on ADE20K \cite{ijcv19/ade20k} $val$ at the resolution of $512^{2}$ using UperNet \cite{eccv18/upernet}. `Conv', `Attn', and `Mamba' denote architectures based on CNN, Transformer, and Mamba, respectively. FLOPs are calculated
using an input size 512$\times$2048. `SS’ and `MS’ denote single-scale and multi-scale testing, respectively. MLN: multi-level neck. $^{\ddagger}$ indicates hierarchical structures while others are plain structures. The \underline{\textbf{best}} results are marked with bold and underline, the \textbf{second-best} with bold only, and the \underline{third-best} with underline only.}
\label{tab:upernet_ade20k_seg}
\begin{tabularx}{\textwidth}{X X}
\resizebox{0.48\textwidth}{!}{
\begin{tabular}{lcccccc}
\toprule
\textbf{Backbone} & \textbf{Type} & \textbf{Params} & \textbf{FLOPs} & \textbf{mIoU (SS)} & \textbf{mIoU (MS)} \\
\midrule
\textbf{ViT-Adpt-T} \cite{iclr23/vit_adpt} & Attn & 36.1 M & - & 42.6 & 43.6 \\
\textbf{Vim-Ti} \cite{icml24/vim} & Mamba & 13.0 M  & - & 41.0 & - \\ 
\textbf{PlainMamba-L1} \cite{bmvc24/plainmamba} & Mamba & 35.0 M & 174.0 G & 44.1 & - \\ 
\textbf{EffVMamba-T} \cite{arxiv24/effvmamba} & Mamba & 14.0 M & 230.0 G & 38.9 & 39.3 \\
\textbf{QuadMamba-T} \cite{neurips24/QuadMamba} & Mamba & 40 M & 886 G & 44.3 & 45.1 \\
\textbf{VSSD-M} \cite{arxiv24/VSSD} & Mamba & 42 M & 893 G & \textbf{45.6} & \textbf{46.0} \\
\midrule
\textbf{ConvNeXt-T} \cite{cvpr22/convnext} & Conv & 60.0 M  & 939.0 G & 46.0 & 46.7 \\
\textbf{MambaOut-T} \cite{arxiv24/mambaout} & Conv & 54.0 M & 938.0 G & 47.4 & 48.6 \\
\textbf{ViT-Adpt-S} \cite{iclr23/vit_adpt} & Attn & 57.6 M & - & 46.2 & 47.1 \\
\textbf{DeiT-S + MLN} \cite{eccv22/revenge_vit} & Attn & 58.0 M  & 1217.0 G & 43.8 & 45.1 \\
\textbf{Swin-T} \cite{iccv21/swinvit} & Attn & 60.0 M  & 945.0 G & 44.4 & 45.8 \\
\textbf{SG-Former-S} \cite{iccv23/sgformer} & Attn & 52.5 M & 989.0 G & \underline{\textbf{49.9}} & \underline{\textbf{51.5}} \\ 
\hdashline
\textbf{Vim-S} \cite{icml24/vim} & Mamba & 46.0 M  & - & 44.9 & - \\ 
\textbf{Mamba$^{\circledR}$-S} \cite{arxiv24/Mamba-R} & Mamba & 56.0 M & - & 45.3 & - \\ 
\textbf{PlainMamba-L2} \cite{bmvc24/plainmamba} & Mamba & 55.0 M & 285.0 G & 46.8 & - \\ 
\textbf{VMamba-T} \cite{neurips24/vmamba} & Mamba & 62.0 M  & 948.0 G & 48.3 & 48.6 \\
\textbf{FractalMamba-T} \cite{arxiv24/FractalMamba} & Mamba & 62.0 M & 948.0 G & \textbf{48.9} & \textbf{49.8} \\
\textbf{EffVMamba-S} \cite{arxiv24/effvmamba} & Mamba & 29.0 M & 505.0 G & 41.5 & 42.1 \\
\textbf{V2M-S$^{\ddagger}$} \cite{arxiv24/v2m} & Mamba & 62 M & - & 48.2 & 49.0 \\
\textbf{GlobalMamba-T$^{\ddagger}$} \cite{arxiv24/GlobalMamba} & Mamba & 62 M & - & 48.1 & 49.0 \\
\textbf{QuadMamba-S} \cite{neurips24/QuadMamba} & Mamba & 62 M & 961 G & 47.2 & 48.1 \\
\textbf{MSVMamba-M} \cite{neurips24/MSVMamba} & Mamba & 42.0 M & 875.0 G & 45.1 & 45.4 \\
\textbf{LocalVim-T} \cite{arxiv24/localmamba} & Mamba & 36.0 M & 181.0 G & 43.4 & 44.4 \\
\textbf{LocalVMamba-T} \cite{arxiv24/localmamba} & Mamba & 57.0 M & 970.0 G & 47.9 & 49.1 \\
\textbf{MambaVision-T} \cite{arxiv24/MambaVision} & Mamba & 55 M & 945 G & 46.6 & - \\
\textbf{GroupMamba-T} \cite{arxiv24/GroupMamba} & Mamba & 49 M & 955 G & \underline{48.6} & 49.2 \\
\textbf{Spatial-Mamba-T} \cite{arxiv24/Spatial-Mamba} & Mamba & 57 M & 936 G & \underline{48.6} & \underline{49.4} \\
\textbf{VSSD-T} \cite{arxiv24/VSSD} & Mamba & 53 M & 941 G & 47.9 & 48.7 \\
\bottomrule
\end{tabular}} &
\resizebox{0.48\textwidth}{!}{
\begin{tabular}{lcccccc}
\toprule
\textbf{Backbone} & \textbf{Type} & \textbf{Params} & \textbf{FLOPs} & \textbf{mIoU (SS)} & \textbf{mIoU (MS)} \\
\midrule
\textbf{ConvNeXt-S} \cite{cvpr22/convnext} & Conv & 82.0 M  & 1027.0 G & 48.7 & 49.6 \\
\textbf{MambaOut-S} \cite{arxiv24/mambaout} & Conv & 76.0 M & 1032.0 G & 49.5 & 50.6 \\
\textbf{Swin-S} \cite{iccv21/swinvit} & Attn & 81.0 M  & 1039.0 G & 47.6 & 49.5 \\
\textbf{SG-Former-M} \cite{iccv23/sgformer} & Attn & 68.3 M & 1114.0 G & \underline{\textbf{51.2}} & \underline{\textbf{52.1}} \\ 
\hdashline
\textbf{PlainMamba-L3} \cite{bmvc24/plainmamba} & Mamba & 81.0 M & 419.0 G & 49.1 & - \\ 
\textbf{VMamba-S} \cite{neurips24/vmamba} & Mamba & 82.0 M  & 1039.0 G & 50.6 & 51.2 \\
\textbf{EffVMamba-B} \cite{arxiv24/effvmamba} & Mamba & 65.0 M & 930.0 G & 46.5 & 47.3 \\
\textbf{V2M-B$^{\ddagger}$} \cite{arxiv24/v2m} & Mamba & 82 M & - & \underline{50.8} & \underline{51.3} \\
\textbf{GlobalMamba-S$^{\ddagger}$} \cite{arxiv24/GlobalMamba} & Mamba & 82 M & - & \textbf{50.9} & \textbf{51.4} \\
\textbf{QuadMamba-B} \cite{neurips24/QuadMamba} & Mamba & 82 M & 1042 G & 49.7 & 50.8 \\
\textbf{MSVMamba-T} \cite{neurips24/MSVMamba} & Mamba & 65.0 M & 942.0 G & 47.6 & 48.5 \\
\textbf{LocalVim-S} \cite{arxiv24/localmamba} & Mamba & 58.0 M & 297.0 G & 46.4 & 47.5 \\
\textbf{LocalVMamba-S} \cite{arxiv24/localmamba} & Mamba & 81.0 M & 1095.0 G & 50.0 & 51.0 \\
\textbf{SiMBA-S} \cite{arxiv24/simba} & Mamba & 62.0 M & 1040.0 G & 49.0 & 49.6 \\
\textbf{MambaVision-S} \cite{arxiv24/MambaVision} & Mamba & 84 M & 1135 G & 48.2 & - \\
\textbf{Spatial-Mamba-S} \cite{arxiv24/Spatial-Mamba} & Mamba & 73 M & 992 G & 50.6 & \textbf{51.4} \\
\midrule
\textbf{ConvNeXt-B} \cite{cvpr22/convnext} & Conv & 122.0 M & 1170.0 G & 49.1 & 49.9 \\
\textbf{MambaOut-B} \cite{arxiv24/mambaout} & Conv & 112.0 M & 1178.0 G & 49.6 & 51.0 \\
\textbf{ViT-Adpt-B} \cite{iclr23/vit_adpt} & Attn & 133.9 M & - & 48.8 & 49.7 \\
\textbf{DeiT-B + MLN} \cite{eccv22/revenge_vit} & Attn & 144.0 M & 2007.0 G & 45.5 & 47.2 \\
\textbf{Swin-B} \cite{iccv21/swinvit} & Attn & 121.0 M & 1188.0 G & 48.1 & 49.7 \\
\textbf{SG-Former-B} \cite{iccv23/sgformer} & Attn & 109.3 M & 1304.0 G & \underline{\textbf{52.0}} & \underline{\textbf{52.7}} \\ 
\hdashline
\textbf{Mamba$^{\circledR}$-B} \cite{arxiv24/Mamba-R} & Mamba & 132.0 M & - & 47.7 & - \\ 
\textbf{Mamba$^{\circledR}$-L} \cite{arxiv24/Mamba-R} & Mamba & 377.0 M & - & 49.1 & - \\ 
\textbf{VMamba-B} \cite{neurips24/vmamba} & Mamba & 122.0 M & 1170.0 G & 51.0 & 51.6 \\
\textbf{GlobalMamba-B$^{\ddagger}$} \cite{arxiv24/GlobalMamba} & Mamba & 122 M & - & \underline{51.2} & \underline{51.7} \\
\textbf{MambaVision-B} \cite{arxiv24/MambaVision} & Mamba & 126 M & 1342 G & 49.1 & - \\
\textbf{Spatial-Mamba-B} \cite{arxiv24/Spatial-Mamba} & Mamba & 127 M & 1176 G & \textbf{51.8} & \textbf{52.6} \\
\bottomrule
\end{tabular}}
\end{tabularx}
\end{table}

\tikzstyle{my-box}=[
    rectangle,
    rounded corners,
    text opacity=1,
    minimum height=1.5em,
    minimum width=5em,
    inner sep=2pt,
    align=center,
    fill opacity=.5,
]
\tikzstyle{my-box-1}=[
    rectangle,
    rounded corners,
    text opacity=1,
    minimum height=1.5em,
    minimum width=15em,
    inner sep=2pt,
    align=center,
    fill opacity=.5,
]
\tikzstyle{cause_leaf}=[my-box, minimum height=1.5em,
    fill=lighttealblue!40, text=black, align=left,font=\scriptsize,
    inner xsep=2pt,
    inner ysep=4pt,
]
\tikzstyle{detect_leaf}=[my-box, minimum height=1.5em, minimum width=26em,
    fill=lightplum!20, text=black, align=left,font=\scriptsize,
    inner xsep=2pt,
    inner ysep=4pt,
]
\tikzstyle{mitigate_leaf}=[my-box, minimum height=1.5em, 
    fill=harvestgold!20, text=black, align=left,font=\scriptsize,
    inner xsep=2pt,
    inner ysep=4pt,
]

\tikzstyle{point_leaf}=[my-box, minimum height=1.5em, minimum width=26em,
    fill=lightblue!20, text=black, align=left,font=\scriptsize,
    inner xsep=2pt,
    inner ysep=4pt,
]

\tikzstyle{modal_leaf}=[my-box, minimum height=1.5em, minimum width=26em,
    fill=lightgreen!20, text=black, align=left,font=\scriptsize,
    inner xsep=2pt,
    inner ysep=4pt,
]

\begin{figure*}[t]
    \centering
    \resizebox{\textwidth}{!}{
        \begin{forest}
            forked edges,
            for tree={
                grow=east,
                reversed=true,
                anchor=base west,
                parent anchor=east,
                child anchor=west,
                base=left,
                font=\small,
                rectangle,
                rounded corners,
                align=left,
                minimum width=4em,
                edge+={darkgray, line width=1pt},
                s sep=3pt,
                inner xsep=2pt,
                inner ysep=3pt,
                ver/.style={rotate=90, child anchor=north, parent anchor=south, anchor=center},
            },
            where level=1{text width=6.5em,font=\scriptsize,}{},
            where level=2{text width=9.5em,font=\scriptsize,}{},
            where level=3{text width=8.0em,font=\scriptsize,}{},
            where level=4{text width=7.5em,font=\scriptsize,}{},
            [
                Application, ver, color=carminepink!100, fill=carminepink!15,
                text=black
                [
                    Image (\S \ref{sec:image}), color=lighttealblue!100, fill=lighttealblue!100, text=black, text width=5.5em
                    [
                        Image Classification (\S \ref{sec:image_classification}), color=lighttealblue!100, fill=lighttealblue!60,  text=black, text width=10em
                        [
                            {\eg~MambaTSR~\cite{neurocomputing/MambaTSR}, MambaMIL~\cite{miccai24/MambaMIL}, Vim4Path~\cite{cvprw/vim4path}, BI-Mamba~\cite{miccai24/BI-Mamba}, MSSM~\cite{kdd24w/MSSM}, RSMamba~\cite{grsl24/Rsmamba_cls}}, color=lighttealblue!100, fill=lighttealblue!40, text=black, text width=34em
                        ]
                    ]
                    [
                        Image Segmentation (\S \ref{sec:image_segmentation}), color=lighttealblue!100, fill=lighttealblue!60, text=black, text width=10em
                        [
                            {\eg~HMNet~\cite{neurips24/HMNet}, ViM-UNet~\cite{midl24s/ViM-UNet}, Swin-UMamba~\cite{miccai24/Swin-umamba}, VM-UNet-V2~\cite{isbra24/VM-UNet-V2}, MM-UNet~\cite{sr24/MM-UNet}, Polyp-Mamba~\cite{miccai24/Polyp-Mamba}, \\ GGVMamba~\cite{miccai24/GGVMamba}, LKM-UNet~\cite{miccai24/LKM-UNet}, SegMamba~\cite{miccai24/SegMamba}, EM-Net~\cite{micccai24/EM-Net}, CAMS-Net~\cite{wacv25/CAMS-Net}, TP-Mamba~\cite{miccai24/TP-Mamba}, PathMamba~\cite{miccai24/PathMamba}, \\ ShapeMamba-EM~\cite{miccai24/ShapeMamba-EM}, Semi-Mamba-UNet~\cite{kbs24/Semi-Mamba-UNet}, RS3Mamba~\cite{grsl24/RS3Mamba}, ChangeMamba~\cite{tgrs24/changemamba}, MambaHSI~\cite{tgrs24/MambaHSI}}, color=lighttealblue!100, fill=lighttealblue!40, text=black, , text width=34em
                        ]
                    ]
                    [
                        Object Detection (\S \ref{sec:image_detection}), color=lighttealblue!100, fill=lighttealblue!60, text=black, text width=10em
                        [
                            {\eg~FER-YOLO-Mamba~\cite{arxiv24/FER-YOLO-Mamba}, VMambaCC~\cite{arxiv24/VMambaCC}, SOAR~\cite{arxiv24/SOAR}}, color=lighttealblue!100, fill=lighttealblue!40, text=black, , text width=34em
                        ]
                    ]
                    [
                        Image Generation (\S \ref{sec:image_generation}), color=lighttealblue!100, fill=lighttealblue!60, text=black, text width=10em
                        [
                                {\eg~ZigMa~\cite{eccv24/Zigma}, VSSM~\cite{icml24w/VSSM}}, color=lighttealblue!100, fill=lighttealblue!40, text=black, , text width=34em
                        ]
                    ] 
                    [
                        Image Restoration (\S \ref{sec:image_restoration}), color=lighttealblue!100, fill=lighttealblue!60, text=black, text width=10em
                        [
                            {\eg~FreqMamba~\cite{mm24/FreqMamba}, MambaIR~\cite{eccv24/MambaIR}, DVMSR~\cite{cvpr24w/DVMSR}, MambaLLIE~\cite{neurips24/MambaLLIE}, ECMamba~\cite{neurips24/ECMamba}, Deform-Mamba~\cite{miccai24/Deform-Mamba}, \\SSUMamba~\cite{tgrs24/SSUMamba}, FMSR~\cite{tmm24/FMSR}, MambaFormerSR~\cite{grsl24/MambaFormerSR}, VMambaSCI~\cite{mm24/vmambasci}}, color=lighttealblue!100, fill=lighttealblue!40, text=black, , text width=34em
                        ]
                    ]
                    [
                        Others (\S \ref{sec:image_others}), color=lighttealblue!100, fill=lighttealblue!60, text=black, text width=10em
                        [
                            {\eg~MambaAD~\cite{neurips24/MambaAD}, MTMamba~\cite{eccv24/MTMamba}, Hamba~\cite{neurips24/Hamba}, SUM~\cite{wacv25/SUM}}, color=lighttealblue!100, fill=lighttealblue!40, text=black, , text width=34em
                        ]
                    ]
                ]
                [
                    Video (\S \ref{sec:video}), color=lightplum!100, fill=lightplum!100, text=black, text width=5.5em
                    [
                        Video Understanding (\S \ref{sec:video_understanding}), color=lightplum!100, fill=lightplum!60, text=black, text width=10em
                        [
                                {\eg~VideoMamba~\cite{eccv24/videomamba}, VideoMamba$\mathbf{^\dag}$~\cite{eccv24/VideoMamba_2}, MambaTrack~\cite{mm24/MambaTrack}, DLT~\cite{mm24/DLT}, VMRNN~\cite{cvpr24w/VMRNN}
                                }
                                    , color=lightplum!100, fill=lightplum!40, text=black, , text width=34em
                        ]
                    ]
                    [
                        Video Generation (\S \ref{sec:video_generation}), color=lightplum!100, fill=lightplum!60, text=black, text width=10em
                        [
                           {\eg~RainMamba~\cite{mm24/RainMamba}, SSMDiff~\cite{iclrw24/SSMDiff}, ZigMa~\cite{eccv24/Zigma}, VFIMamba~\cite{neurips/VFIMamba}, MMD~\cite{mm24/MMD}}
                                , color=lightplum!100, fill=lightplum!40, text=black, text width=34em
                        ]
                    ]
                ]
                [
                    Point Cloud (\S \ref{sec:point_cloud}), color=lightblue!100, fill=lightblue!100, text=black, text width=5.5em
                            [
                                {\eg~PointMamba~\cite{neurips24/PointMamba}, Mamba3D~\cite{mm24/Mamba3D}, 3DMambaIPF~\cite{3dmambaipf}, LCM~\cite{3dmambacomplete},
                                VoxelMamba~\cite{neurips24/voxelmamba}, MambaMOS~\cite{mm24/MambaMOS}
                                }, color=lightblue!100, fill=lightblue!40, text=black, text width=34em
                            ]
                ]
                [
                    Multi-modal (\S \ref{sec:mm}), color=lightgreen!100, fill=lightgreen!100, text=black, text width=5.5em
                    [
                        Homogeneous Multi-modal (\S \ref{Homogeneous}), color=lightgreen!100, fill=lightgreen!60, text=black, text width=10em
                        [
                       {\eg~MGMF~\cite{tgrs24/MGMF},  Sigma~\cite{wacv25/sigma}
                       }
                                , color=lightgreen!100, fill=lightgreen!40, text=black, text width=34em
                        ]
                    ]
                    [
                        Heterogeneous Multi-modal (\S \ref{Heterogeneous}), color=lightgreen!100, fill=lightgreen!60, text=black, text width=10em
                        [
                                {\eg~InstructGIE~\cite{eccv24/InstructGIE}, Meteor~\cite{neurips24/Meteor}, RoboMamba~\cite{neurips24/RoboMamba}, ReMamber~\cite{eccv24/remamber}, RSCaMa~\cite{grsl24/RSCaMa}, CoupledMamba~\cite{neurips24/Coupled_Mamba}, \\MotionMamba~\cite{eccv24/MotionMamba},
                                MambaTalk~\cite{neurips24/mambatalk},
                                MambaGesture~\cite{mm24/MambaGesture}
                                }
                                , color=lightgreen!100, fill=lightgreen!40, text=black, text width=34em
                        ]
                    ]
                ]
            ]
        \end{forest}
    }
    \caption{The main content flow and categorization of the application in this survey.}
    \label{fig:mamba_app}
\end{figure*}

\section{Applications}
\label{sec:applications}
In this section, we systematically categorize and discuss the diverse vision applications of Mamba. The categorization scheme, along with an overview of the reviewed literature in this survey, is presented in Fig.~\ref{fig:mamba_app}.

\subsection{Image}
\label{sec:image}
The processing approaches used in image applications are built upon the techniques employed in visual Mamba backbone networks, including tokenization methods, scanning strategies, and block designs, among others.

\subsubsection{Classification}
\label{sec:image_classification}
In addition to backbones like Vim~\cite{icml24/vim} and VMamba~\cite{neurips24/vmamba}, which are primarily used for image classification in representation learning, the efficiency and scalability of Mamba-based architectures in handling long sequences have enabled their application across a broad range of image classification tasks~\cite{arxiv24/InsectMamba,neurocomputing/MambaTSR,bibm24/MamMIL,miccai24/MambaMIL,cvprw/vim4path,CMViM,arxiv24/nnMamba,MedMamba,miccai24/BI-Mamba,kdd24w/MSSM,grsl24/Rsmamba_cls}. This makes them particularly effective for high-dimensional image analysis, (\textit{e.g.}, Whole Slide Images, 3D Medical Images, and Remote Sensing Images). 

MambaTSR~\cite{neurocomputing/MambaTSR} leveraged EfficientVMamba~\cite{arxiv24/effvmamba} blocks to achieve efficient and accurate traffic sign recognition.

With regards to high-dimensional image analysis, MambaMIL~\cite{miccai24/MambaMIL} utilized Mamba to enhance Multiple Instance Learning (MIL) for histopathology image analysis. 
It introduced a new approach named Sequence Reordering Mamba (SR-Mamba), which is sensitive to the order and distribution of instances and leverages the valuable information embedded within these long sequences. 
Differently, Vim4Path~\cite{cvprw/vim4path} utilized Vim architecture within the DINO~\cite{iccv21/dino} framework for representation learning. 
BI-Mamba~\cite{miccai24/BI-Mamba} employed parallel forward and backward Mamba blocks to effectively capture long-range dependencies in multi-view high-resolution chest X-rays. MSSM~\cite{kdd24w/MSSM} introduced a multi-scale SSM module for 4D rs-fMRI data, specifically applied to the classification of major depressive disorder.

In the remote sensing scenario, RSMamba~\cite{grsl24/Rsmamba_cls} introduced a position-sensitive dynamic multi-path activation mechanism, encompassing forward, backward, and shuffle paths, to extract precise semantic cues for accurate scene discrimination. 


\begin{table}[!t]
\label{tab:image}
\centering
\caption{The Mamba-based methods in image modality vision tasks. 
The abbreviations of Data here are 
CXR: Chest X-Ray, 
CT: Computed Tomography, 
MRI: Magnetic Resonance Imaging, 
rs-fMRI: Resting-state Functional Magnetic Resonance Imaging, 
EM: Electron Microscopy, 
Echo: Echocardiograms, 
Micro: Microscopy Images, 
Skin: Skin Lesion Images, 
Path: Pathology. 
The abbreviations of Task here are CLS: Classification, DET: Detection, SEG: Segmentation, 
SU: Scene Understanding. 
Tok: Tokenization.}


\resizebox{\linewidth}{!}{
\begin{tabular}{lccccccccc}
\toprule
\multirow{2}{*}{\textbf{Methods}} 
& \multirow{2}{*}{\textbf{Data}}
& \multirow{2}{*}{\textbf{Task}}
& \multirow{2}{*}{\textbf{Tok.}}
& \multicolumn{4}{c}{\textbf{Scan Strategy}}                       
& \multirow{2}{*}{\textbf{Venue}}
& \multirow{2}{*}{\textbf{Code}}
\\ \cmidrule{5-8}
                  
&  
&
&
& \textbf{Direction}
& \textbf{Axis}
& \textbf{Continuity}
& \textbf{Sampling} 
& 
& 
\\
\midrule
\rowcolor{dino}
\multicolumn{10}{c}{\textbf{Image Classification}} \\
\midrule
\textbf{MambaTSR}~\cite{neurocomputing/MambaTSR}
& Natural Images
& Traffic Sign CLS
& 2D
& BD 
& H/V
& Raster
& Global
& Neurocomputing24
& \href{https://github.com/1024AILab/MambaTSR}{\color{magenta}\Checkmark}
\\
\textbf{MambaMIL}~\cite{miccai24/MambaMIL}
& Whole Slide Images 
& Cancer Subtyping/Survival Prediction
& 1D
& SD 
& H
& Raster + Reordering
& Global
& MICCAI24
& \href{https://github.com/isyangshu/MambaMIL}{\color{magenta}\Checkmark}
\\
\textbf{Vim4Path}~\cite{cvprw/vim4path}
& Whole Slide Images
& Cancer Subtyping
& 1D
& BD 
& H
& Raster
& Global
& CVPR24 Workshop
& \href{https://github.com/AtlasAnalyticsLab/Vim4Path}{\color{magenta}\Checkmark}
\\
\textbf{BI-Mamba}~\cite{miccai24/BI-Mamba}
& 2D Medical Images (X-ray)
& Risk Prediction
& 1D
& BD 
& H
& Raster
& Global
& MICCAI24 Oral
& \href{https://github.com/RPIDIAL/BI-Mamba}{\color{magenta}\Checkmark}
\\
\textbf{MSSM}~\cite{kdd24w/MSSM}
& 4D Medical Images (rs-fMRI)
& Major Depressive Disorder CLS
& 1D
& SD 
& H
& Raster
& Global
& KDD24 Workshop
& 
\\
\textbf{RSMamba}~\cite{grsl24/Rsmamba_cls}
& Remote Sensing Images
& Remote Sensing Images CLS
& 1D
& BD 
& H/V
& Raster + Shuffle
& Global
& GRSL24
& \href{https://github.com/KyanChen/RSMamba}{\color{magenta}\Checkmark}
\\
\midrule
\rowcolor{dino}
\multicolumn{10}{c}{\textbf{Image Segmentation}} \\
\midrule
\textbf{HMNet}~\cite{neurips24/HMNet}
& Natural Images
& Few-shot SEG
& 2D
& BD 
& H/W
& Raster
& Global
& NeurIPS24
& \href{https://github.com/Sam1224/HMNet}{\color{magenta}\Checkmark}
\\
\textbf{ViM-UNet}~\cite{midl24s/ViM-UNet}
& 2D Medical Images (Cell/Neurite)
& 2D Medical Images SEG
& 1D
& BD 
& H
& Raster
& Global
& MIDL24 Short 
& \href{https://github.com/constantinpape/torch-em/blob/main/vimunet.md}{\color{magenta}\Checkmark} 
\\
\textbf{Swin-UMamba}~\cite{miccai24/Swin-umamba}
& 2D Medical Images (Endo/Micro/MRI)
& 2D Medical Image SEG
& 2D
& BD 
& H/V
& Raster
& Global
& MICCAI24
& \href{https://github.com/JiarunLiu/Swin-UMamba}{\color{magenta}\Checkmark}
\\
\textbf{VM-UNet-V2}~\cite{isbra24/VM-UNet-V2}
& 2D Medical Images (Skin/Endo)
& 2D Medical Image SEG
& 2D
& BD 
& H/V
& Raster
& Global
& ISBRA24
& \href{https://github.com/nobodyplayer1/VM-UNetV2}{\color{magenta}\Checkmark}
\\
\textbf{MM-UNet}~\cite{sr24/MM-UNet}
& 2D/3D Medical Images (MRI)
& 2D/3D Medical Image SEG
& 1D
& SD 
& H
& Raster
& Global
& Scientific Reports24
& 
\\
\textbf{Polyp-Mamba}~\cite{miccai24/Polyp-Mamba}
& 2D Medical Images (Endo)
& 2D Medical Image SEG
& 2D
& BD 
& H/V
& Raster
& Global
& MICCAI24
& 
\\
\textbf{GGVMamba}~\cite{miccai24/GGVMamba}
& 2D Medical Images (CXR)
& 2D Medical Image SEG
& 1D/2D
& BD 
& H/V
& Raster
& Global
& MICCAI24 
& \href{https://github.com/SCU-zly/GGVMamba}{\color{magenta}\Checkmark}
\\
\textbf{LKM-UNet}~\cite{miccai24/LKM-UNet}
& 2D/3D Medical Images (CT/MRI)
& 2D/3D Medical Image SEG
& 1D
& BD 
& H
& Raster
& Global/Local
& MICCAI24 
& \href{https://github.com/wjh892521292/LKM-UNet}{\color{magenta}\Checkmark}
\\
\textbf{SegMamba}~\cite{miccai24/SegMamba}
& 3D Medical Images (CT/MRI)
& 3D Medical Image SEG
& 1D
& BD + SD
& H + D
& Raster
& Global
& MICCAI24
& \href{https://github.com/ge-xing/SegMamba}{\color{magenta}\Checkmark}
\\
\textbf{EM-Net}~\cite{micccai24/EM-Net}
& 3D Medical Images (CT)
& 3D Medical Image SEG
& 1D
& SD 
& C
& Raster
& Global
& MICCAI24 
& \href{https://github.com/zang0902/EM-Net}{\color{magenta}\Checkmark}
\\
\textbf{CAMS-Net}~\cite{wacv25/CAMS-Net}
& 2D Medical Images (MRI)
& 2D Medical Image SEG
& 1D
& BD 
& H/C
& Raster
& Global
& WACV25 
& \href{https://github.com/kabbas570/CAMS-Net}{\color{magenta}\Checkmark}
\\
\textbf{TP-Mamba}~\cite{miccai24/TP-Mamba}
& 3D Medical Images (CT)
& 3D Medical Image SEG
& 1D
& SD 
& H/V/D
& Raster
& Global
& MICCAI24 
& \href{https://github.com/xmed-lab/TP-Mamba}{\color{magenta}\Checkmark}
\\
\textbf{ShapeMamba-EM}~\cite{miccai24/ShapeMamba-EM}
& 3D Medical Images (EM)
& 3D Medical Image SEG
& 1D
& SD 
& H
& Raster
& Global
& MICCAI24
& 
\\
\textbf{PathMamba}~\cite{miccai24/PathMamba}
& 2D Medical Images (Path)
& 2D Medical Image SEG
& 1D
& BD 
& H
& Raster
& Global
& MICCAI24 
& \href{https://github.com/hemo0826/PathMamba}{\color{magenta}\Checkmark}
\\
\textbf{Semi-Mamba-UNet}~\cite{kbs24/Semi-Mamba-UNet}
& 2D Medical Images (MRI)
& 2D Medical Image SEG
& 2D
& BD 
& H/V
& Raster
& Global
& KBS24
& \href{https://github.com/ziyangwang007/Mamba-UNet}{\color{magenta}\Checkmark}
\\
\textbf{RS3Mamba}~\cite{grsl24/RS3Mamba}
& Remote Sensing Images
& Semantic SEG
& 2D
& BD 
& H/V
& Raster
& Global
& GRSL24
& \href{https://github.com/sstary/SSRS}{\color{magenta}\Checkmark}
\\
\textbf{ChangeMamba}~\cite{tgrs24/changemamba}
& Remote Sensing Images
& Change DET/Building Damage Assessment
& 2D
& BD 
& H/V
& Raster
& Global
& TGRS24
& \href{https://github.com/ChenHongruixuan/MambaCD}{\color{magenta}\Checkmark}
\\
\textbf{MambaHSI}~\cite{tgrs24/MambaHSI}
& Remote Sensing Images
& Remote Sensing Images CLS
& 1D
& SD 
& H + C
& Raster
& Global
& TGRS24
& \href{https://github.com/li-yapeng/MambaHSI}{\color{magenta}\Checkmark}
\\
\midrule
\rowcolor{dino}
\multicolumn{10}{c}{\textbf{Image Generation}} \\
\midrule
\textbf{ZigMa}~\cite{eccv24/Zigma}
& Natural Images
& Image Generation
& 1D
& BD 
& H/V
& Zigzag
& Global
& ECCV24 
& \href{https://github.com/CompVis/zigma}{\color{magenta}\Checkmark}
\\
\textbf{VSSM}~\cite{icml24w/VSSM}
& Natural Images
& Image Generation
& 1D
& SD 
& H
& Raster
& Global
& ICML24 Workshop
& 
\\
\midrule
\rowcolor{dino}
\multicolumn{10}{c}{\textbf{Image Restoration}} \\
\midrule
\textbf{FreqMamba}~\cite{mm24/FreqMamba}
& Natural Images
& Deraining
& 2D
& BD
& H/V + F
& Raster
& Global/Local
& MM24
& \href{https://github.com/aSleepyTree/FreqMamba}{\color{magenta}\Checkmark}
\\
\textbf{MambaIR}~\cite{eccv24/MambaIR}
& Natural Images
& Super-resolution/Denoising
& 2D
& BD 
& H/V
& Raster
& Global
& ECCV24
& \href{https://github.com/csguoh/MambaIR}{\color{magenta}\Checkmark}
\\
\textbf{DVMSR}~\cite{cvpr24w/DVMSR}
& Natural Images
& Super-resolution
& 1D
& SD 
& H
& Raster
& Global
& CVPR24 Workshop
& \href{https://github.com/nathan66666/DVMSR}{\color{magenta}\Checkmark}
\\
\textbf{MambaLLIE}~\cite{neurips24/MambaLLIE}
& Natural Images
& Enhancement
& 2D
& BD 
& H/V
& Raster
& Global
& NeurIPS24
& 
\\
\textbf{ECMamba}~\cite{neurips24/ECMamba}
& Natural Images
& Enhancement
& 2D
& BD 
& H/V
& Reordering
& Global
& NeurIPS24
& \href{https://github.com/LowlevelAI/ECMamba}{\color{magenta}\Checkmark}
\\
\textbf{Deform-Mamba}~\cite{miccai24/Deform-Mamba}
& 2D Medical Images (MRI)
& Super-resolution
& 2D
& BD 
& H/V
& Raster
& Global
& MICCAI24
& 
\\
\textbf{SSUMamba}~\cite{tgrs24/SSUMamba}
& Remote Sensing Images
& Hyperspectral Denoising
& 1D
& BD 
& H/V/D 
& Raster
& Global
& TGRS24
& \href{https://github.com/lronkitty/SSUMamba}{\color{magenta}\Checkmark}
\\
\textbf{FMSR}~\cite{tmm24/FMSR}
& Remote Sensing Images
& Super-resolution
& 2D
& BD 
& H/V
& Raster
& Global
& TMM24
& \href{https://github.com/XY-boy/FreMamba}{\color{magenta}\Checkmark}
\\
\textbf{MambaFormerSR}~\cite{grsl24/MambaFormerSR}
& Remote Sensing Images
& Super-resolution
& 2D
& BD 
& H/V
& Raster
& Global
& GRSL24
& 
\\
\textbf{VMambaSCI}~\cite{mm24/vmambasci}
& Remote Sensing Images
& Spectral Reconstruction
& 2D
& BD 
& H/V + C
& Raster
& Global
& MM24
& 
\\
\midrule
\rowcolor{dino}
\multicolumn{10}{c}{\textbf{Others}} \\
\midrule
\textbf{MambaAD}~\cite{neurips24/MambaAD}
& Natural Images
& Multi-class Anomaly DET
& 2D
& BD 
& H/V/H*/V* 
& Hilbert
& Global
& NeurIPS24
& \href{https://github.com/lewandofskee/MambaAD}{\color{magenta}\Checkmark}
\\
\textbf{MTMamba}~\cite{eccv24/MTMamba}
& Natural Images
& Multi-task Dense SU
& 2D
& BD 
& H/V
& Raster
& Global
& ECCV24
& \href{https://github.com/EnVision-Research/MTMamba}{\color{magenta}\Checkmark}
\\
\textbf{Hamba}~\cite{neurips24/Hamba}
& Natural Images
& 3D Hand Reconstruction
& 2D
& BD 
& Graph-guided
& Raster
& Global
& NeurIPS24 
& \href{https://github.com/humansensinglab/Hamba}{\color{magenta}\Checkmark}
\\
\textbf{SUM}~\cite{wacv25/SUM}
& Natural Images
& Visual Attention Modeling
& 2D
& BD 
& H/V
& Raster
& Global
& WACV25 
& \href{https://github.com/Arhosseini77/SUM}{\color{magenta}\Checkmark}
\\
\bottomrule
\end{tabular}
}
\end{table}

\subsubsection{Segmentation}
\label{sec:image_segmentation}
Segmentation remains a vital and prominent area in computer vision, holding immense value for diverse real-world applications~\cite{MIM-ISTD,sr24/MM-UNet,U-Mamba,miccai24/SegMamba,Vm-unet,miccai24/Swin-umamba,arxiv24/nnMamba,Mamba-UNet,kbs24/Semi-Mamba-UNet,P-Mamba,Weak-Mamba-Unet,LightM-UNet,LW-Mamba-UNet,isbra24/VM-UNet-V2,H-vmunet,miccai24/LKM-UNet,miccai24/Promamba,RTS,IMH,ultralightvmunet,TMamba,midl24s/ViM-UNet,miccai24/nnU-Net_revisit,arxiv24/Mamba-Ahnet,arxiv24/AC-MAMBASEG,arxiv24/HC-Mamba,arxiv24/MUCMNet,arxiv24/UU-Mamba,arxiv24/TokenUnify,miccai24/ShapeMamba-EM,miccai24/TP-Mamba,micccai24/EM-Net,miccai24/PathMamba,miccai24/GGVMamba,Samba,grsl24/RS3Mamba,yao2024spectralmamba,wang2024s,arxiv24/SS-Mamba,arxiv24/rethinking,liu2024cm,RSMamba,tgrs24/changemamba,zhou2024mamba,he20243dss,tgrs24/MambaHSI,bmvc24/medical_seg_robustness,accv24/OneBEV,wacv25/CAMS-Net}. Mamba is well-positioned to enhance segmentation tasks for its strong long-range modeling capability.

In the context of few-shot segmentation, HMNet~\cite{neurips24/HMNet} identified two issues when concatenating support and query features and applying the original Mamba: support forgetting and the intra-class gap between support and query. To mitigate these issues, HMNet introduced two mechanisms: periodically re-scanning the support features during the query feature scanning process, and scanning the query features in parallel to avoid undesired pixel-level interactions within the query features. 

Recent advances based on Mamba have appeared in the medical domain, encompassing diverse 2D and 3D medical image datasets~\cite{arxiv24/UU-Mamba,arxiv24/TokenUnify}. Several methods~\cite{midl24s/ViM-UNet,miccai24/Swin-umamba,isbra24/VM-UNet-V2} directly replace CNN-blocks in the U-Net architecture with Mamba-based blocks~\cite{arxiv23/mamba,neurips24/vmamba,icml24/vim}. U-Mamba~\cite{U-Mamba} is the first effective attempt to apply Mamba in medical image segmentation. It introduced a hybrid CNN-SSM block. ViM-UNet \cite{midl24s/ViM-UNet} utilized Vim~\cite{icml24/vim} block as the basic blocks. Swin-UMamba~\cite{miccai24/Swin-umamba} and VM-UNet-V2~\cite{isbra24/VM-UNet-V2} employed the VMamba~\cite{neurips24/vmamba} block as a basic block to capture extensive contextual information. MM-UNet~\cite{sr24/MM-UNet}, Polyp-Mamba~\cite{miccai24/Polyp-Mamba}, GGVMamba~\cite{miccai24/GGVMamba}, and LKM-UNet~\cite{miccai24/LKM-UNet} employed Mamba-based architecture as a basic block with additional designs for enhancement. 
SegMamba~\cite{miccai24/SegMamba} introduced gated spatial convolution and tri-orientated Spatial Mamba, which utilized an inter-slice path for 3D data. EM-Net~\cite{micccai24/EM-Net} employed a channel scanning strategy and integrated frequency analysis. CAMS-Net~\cite{wacv25/CAMS-Net} introduced Mamba-based channel and spatial aggregators. 
TP-Mamba~\cite{miccai24/TP-Mamba} and ShapeMamba-EM~\cite{miccai24/ShapeMamba-EM} each introduced a 3D adapter based on Mamba to fine-tune the foundation models SAM~\cite{iccv23/SAM} and SAM-Med3D~\cite{arxiv24/SAM-Med3D}, respectively, for medical image data. 
PathMamba~\cite{miccai24/PathMamba} incorporated Mamba into weakly supervised multi-label segmentation of pathology images and introduced a dual-granularity comparative Mamba block, designed to facilitate pixel- and patch-level contrastive learning. In \cite{kbs24/Semi-Mamba-UNet}, a purely Mamba-based UNet and a CNN-based UNet were integrated into a semi-supervised learning framework, where they perform cross-supervision and contrastive learning.

In the field of remote sensing images, Mamba-based approaches typically leverage its efficiency for processing high-resolution images~\cite{grsl24/RS3Mamba}, and exploit the spatial-temporal and spatial-spectral characteristics of images to perform scanning~\cite{tgrs24/changemamba,tgrs24/MambaHSI}. 

\subsubsection{Detection}
\label{sec:image_detection}
The advantages of Mamba in extracting global information and performing efficient computations make it particularly well-suited for object detection tasks. 

FER-YOLO-Mamba~\cite{arxiv24/FER-YOLO-Mamba} is the first Mamba-based model for facial expression detection. It introduced a dual-branch structure that incorporated omnidirectional Mamba and attention mechanisms for the feature pyramid network. VMambaCC~\cite{arxiv24/VMambaCC} is a pioneering work that applied the VMamba~\cite{neurips24/vmamba} architecture to crowd counting tasks. SOAR~\cite{arxiv24/SOAR} integrated vim~\cite{icml24/vim} block into the YOLO v9 architecture.

\subsubsection{Generation}
\label{sec:image_generation}
Intuitively, applying the Mamba architecture to a series of generation tasks to achieve sufficient long sequence interactions has the potential to achieve impressive performance~\cite{DiS,eccv24/Zigma,arxiv24/DiM,arxiv24/DiM1,icml24w/VSSM,Gamba,MD-Dose,arxiv24/VM-DDPM,arxiv24/SMCD}.

ZigMa~\cite{eccv24/Zigma} is a DiT-style~\cite{iccv23/dit} Mamba diffusion model. It leveraged spatial continuity to maximally incorporate the inductive bias from visual data. VSSM~\cite{icml24w/VSSM} is a variational autoencoder in which both the encoder and decoder are SSMs, enabling efficient parallel computation.

\subsubsection{Restoration}
\label{sec:image_restoration}
Recently, the Mamba architecture has also been extensively applied to various image restoration tasks. These low-level tasks include reconstructing degraded images to their original state, such as image dehazing~\cite{UVMNet,arxiv24/RSDehamba}, deraining~\cite{UVMNet,mm24/FreqMamba,arxiv24/DFSSM,arxiv24/FourierMamba,VmambaIR}, denoising~\cite{arxiv24/CU-Mamba,eccv24/MambaIR,arxiv24/HSIDMamba,tgrs24/SSUMamba}, deblurring~\cite{arxiv24/CU-Mamba,arxiv24/ALGNet,arxiv24/EVSSM}, super-resolution~\cite{eccv24/MambaIR,VmambaIR,MMA,cvpr24w/DVMSR,arxiv24/IRSRMamba,miccai24/Deform-Mamba,tmm24/FMSR,grsl24/MambaFormerSR}, and other reconstruction~\cite{FDVMNet,MambaMIR,arxiv24/MambaMIR,arxiv24/MSDR,arxiv24/GMSR,mm24/vmambasci}, and enhancing the image quality by adjusting luminance, contrast, and other visual attributes, such as multiple exposure correction~\cite{accv24/wavelet}, low light enhancement~\cite{UVMNet,neurips24/MambaLLIE,arxiv24/Retinexmamba}, and underwater image enhancement~\cite{MambaUIE,arxiv24/WaterMamba}. 

The extraction of local fine-grained features plays a critical role in tackling low-level image tasks, which the vanilla Mamba architecture lacks. Besides, image restoration approaches often utilize techniques from the frequency domain, leveraging the ability to isolate and manipulate specific image frequency components for improving detail recovery. For instance, FreqMamba~\cite{mm24/FreqMamba} utilized both Global and Local scanning, effectively removing high-frequency rain streaks in deraining tasks by leveraging frequency-based techniques. 
Similarly, in image super-resolution tasks, MambaIR~\cite{eccv24/MambaIR} incorporated convolutions and channel attentions to enhance the Mamba architecture's capacity for preserving local information while effectively reducing channel redundancy. Differently, DVMSR~\cite{cvpr24w/DVMSR} applied a distillation strategy to the Mamba-based network to enhance its efficiency. 
In image enhancement, MambaLLIE~\cite{neurips24/MambaLLIE} enhanced local dependencies by incorporating an augmented local bias within the 2D selective scan mechanism and introducing an implicit Retinex-aware selective kernel module. ECMamba~\cite{neurips24/ECMamba} integrated Retinex theory for exposure correction and presented a feature-aware scanning strategy that leverages an activation response map from deformable convolution to generate sequences ordered by feature importance.

In the medical domain, Deform-Mamba~\cite{miccai24/Deform-Mamba} combined the modulated deformable block with VMamba~\cite{neurips24/vmamba} block to effectively capture both local and global information. Additionally, it introduced a multi-view context module and a contrastive edge loss.

In the remote sensing domain, to improve computational efficiency in hyperspectral denoising, SSUMamba~\cite{tgrs24/SSUMamba} introduced a spatial-spectral continuous scan Mamba, alternating rows, columns, and bands in six different orders to capture long-range spatial-spectral dependencies. Furthermore, 3D convolutions are incorporated to enhance local spatial-spectral modeling. In remote sensing image super-resolution tasks, FMSR~\cite{tmm24/FMSR} enhanced spatial-frequency fusion by integrating frequency selection modules with the global VMamba~\cite{neurips24/vmamba} block and a local hybrid gating module. MambaFormerSR~\cite{grsl24/MambaFormerSR} combined Mamba with Transformer architectures, while also introducing a convolutional Fourier transform feed-forward network. VMambaSCI~\cite{mm24/vmambasci} introduced a novel spatial-channel scanning technique for spectral reconstruction.

\subsubsection{Others}
\sloppy
\label{sec:image_others}
Mamba's ability to model long-range dependencies extends its advantages to various other image modality tasks~\cite{neurips24/MambaAD,arxiv24/MambaVC,eccv24/MTMamba,neurips24/Hamba,wacv25/SUM}, improving feature representation and contextual understanding across diverse applications. MambaAD~\cite{neurips24/MambaAD} applied Mamba to multi-class unsupervised anomaly detection by combining a pre-trained encoder with a Mamba-based decoder. The decoder integrated parallel cascaded hybrid state space blocks, utilizing Hilbert scanning in eight directions, and multi-kernel convolution operations at multiple scales to effectively capture both global and local information. MTMamba~\cite{eccv24/MTMamba} introduced self-task and cross-task Mamba blocks for multi-task dense prediction. In particular, the cross-task Mamba block was designed to enhance interactions between tasks. Hamba~\cite{neurips24/Hamba} introduced a graph-guided bidirectional scanning technique for 3D hand reconstruction from a single RGB image, efficiently utilizing a few effective tokens to learn the joint relations and spatial sequences. SUM~\cite{wacv25/SUM} leveraged Mamba for efficient long-range dependency modeling and introduced a novel conditional visual state space block, enabling dynamic adaptation to different image types, in visual attention modeling tasks.

\begin{table}[h]
\label{tab:video}
\centering
\caption{The Mamba-based methods in video modality vision tasks. 
The abbreviations of Task here are 
CLS: Classification, 
VU: Video Understanding, 
T2VR: Text-to-Video Retrieval, 
MOT: Multi-object Tracking, 
STF: Spatiotemporal Forecasting,
VFI: Video Frame Interpolation.}
\resizebox{0.8\linewidth}{!}
{
\begin{tabular}{lcccccc}
\toprule
\textbf{Methods}
& \textbf{Data}
& \textbf{Task}
& \textbf{Scan Strategy}
& \textbf{Venue}
& \textbf{Code}
\\
\midrule
\rowcolor{dino}
\multicolumn{6}{c}{\textbf{Video Understanding}} \\
\midrule
\textbf{VideoMamba}~\cite{eccv24/videomamba}
& Natural Videos
& Action CLS, VU, T2VR
& BD Spatial-first
& ECCV24
&\href{https://github.com/OpenGVLab/VideoMamba}{\color{magenta}\Checkmark}
\\
\textbf{VideoMamba}$\mathbf{^\dag}$~\cite{eccv24/VideoMamba_2}
& Natural Videos
& VU
& BD Spatial-first
& ECCV24
&\href{https://github.com/jnypark/VideoMamba}{\color{magenta}\Checkmark}
\\
\textbf{MambaTrack}~\cite{mm24/MambaTrack}
& Natural Videos
& MOT
& BD Spatial-first
& MM24 Oral
&
\\
\textbf{DLT}~\cite{mm24/DLT}
& Natural Videos
& MOT
& BD Spatial-first
& MM24
&
\\
\textbf{VMRNN}~\cite{cvpr24w/VMRNN}
& Natural Videos
& STF
& Spatial
& CVPR24 Workshop
& \href{https://github.com/yyyujintang/VMRNN-PyTorch}{\color{magenta}\Checkmark}
\\
\midrule
\rowcolor{dino}
\multicolumn{6}{c}{\textbf{Video Generation}} \\
\midrule
\textbf{RainMamba}~\cite{mm24/RainMamba}
& Natural Videos
& Video Deraining
& BD Spatial-first (Zigzag) + 3D Hilbert
& MM24 Oral
&\href{https://github.com/TonyHongtaoWu/RainMamba}{\color{magenta}\Checkmark}
\\
\textbf{SSMDiff}~\cite{iclrw24/SSMDiff}
& Natural Videos
& Video Generation
& Spatial + Temporal
& ICLR24 Workshop
& \href{https://github.com/shim0114/SSM-Meets-Video-Diffusion-Models}{\color{magenta}\Checkmark}\\
\textbf{ZigMa}~\cite{eccv24/Zigma}
& Natural Videos
& Video Generation
& Spatial + Temporal
& ECCV24
& \href{https://github.com/CompVis/zigma}{\color{magenta}\Checkmark}\\
\textbf{VFIMamba}~\cite{neurips/VFIMamba}
& Natural Videos
& VFI
& Temporal-first
& NeurIPS24
& \href{https://github.com/MCG-NJU/VFIMamba}
{\color{magenta}\Checkmark}\\
\textbf{MMD}~\cite{mm24/MMD}
& Motion Sequence
& Motion Generation
& Spatial + Temporal
& MM24
& 
\\
\bottomrule
\end{tabular}
}
\end{table}

\begin{wrapfigure}{r}{0.42\textwidth}
  \centering
  \includegraphics[width=0.4\textwidth]{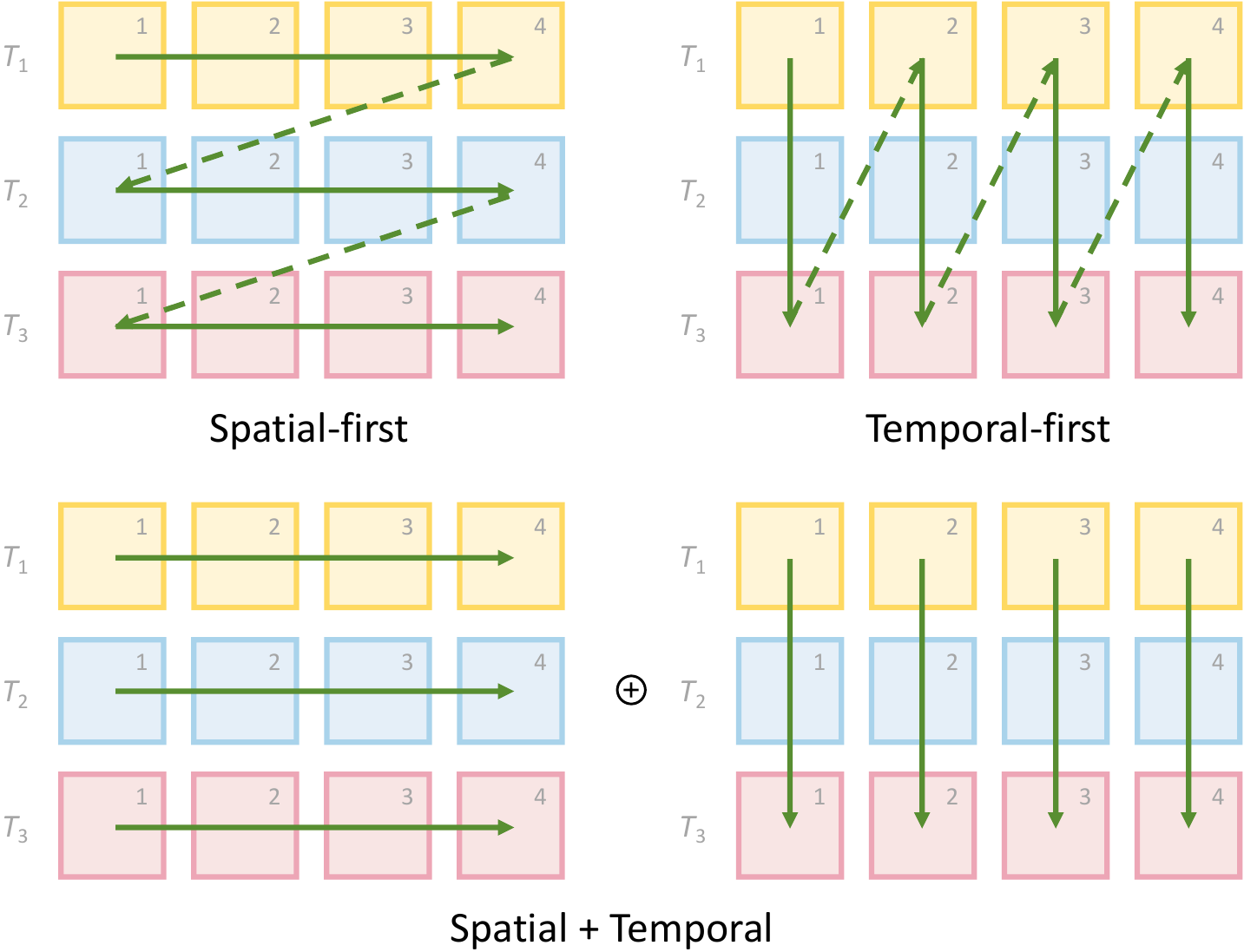}
  \caption{Video scanning techniques.}
  \Description{}
  \label{fig:video_scan}
\end{wrapfigure}

\subsection{Video}
\label{sec:video}
Video processing is a key area in computer vision, dealing with 3D video data that consists of sequences of 2D frames. Video data includes a temporal dimension in addition to the spatial dimensions, enabling the capture of temporal dynamics. The primary goal of video processing is to effectively master spatio-temporal representations across long contexts. Mamba excels in this domain with its selective state space model, which achieves a balance between maintaining linear complexity and enabling effective long-term dynamic modeling. To adapt Mamba to 3D video data, current approaches organize the 3D video tokens in the following ways. 

\textbf{Spatial-first} scans spatial tokens by location and then stacks them frame by frame. The scanning techniques for spatial tokens within each frame are similar to those used for 2D images. \textbf{Temporal-first} arranges temporal tokens by frame and then stacks them along the spatial dimension. To improve the receptive field and mitigate Mamba's position sensitivity, these spatiotemporal scanning paths can be directly reversed. Considering the sequential attribute of the temporal dimension, Spatial Reversal reverses the spatial tokens in each frame and maintains the order along the temporal axis. Temporal Reversal maintains the order of spatial tokens and reverses the temporal sequence. 

Additionally, some approaches apply Mamba separately to spatial and temporal information, utilizing 2D \textbf{Spatial} scanning and 1D \textbf{Temporal} scanning. Other approaches focus solely on Spatial scanning, leveraging alternative techniques to handle temporal dynamics. The video data can also be regarded as 3D data. In this situation, scanning techniques like 3D zigzag scanning and 3D Hilbert scanning can be utilized. 

These innovative approaches, illustrated in Fig.~\ref{fig:video_scan}, have been widely adopted in diverse video analysis tasks such as video understanding~\cite{eccv24/videomamba} and video generation~\cite{eccv24/Zigma}.

\subsubsection{Video Understanding}
\label{sec:video_understanding}
Recent approaches in video understanding tasks have adopted Mamba as the core architecture due to its superior long-range modeling capability and computational efficiency for analyzing high-resolution long video data, with a primary focus on optimizing scanning techniques~\cite{vivim,eccv24/videomamba,eccv24/VideoMamba_2,videomambasuite,rhythmmamba,mm24/MambaTrack,mm24/DLT,cvpr24w/VMRNN,arxiv24/DeMamba,arxiv24/MAMBA4D}. 
Both VideoMamba~\cite{eccv24/videomamba} and VideoMamba$\mathbf{^\dag}$~\cite{eccv24/VideoMamba_2} built a purely Mamba-based model by stacking identical bi-directional Mamba blocks and utilizing the BD Spatial-first scanning strategy. VideoMamba~\cite{eccv24/videomamba} conducted extensive experiments comparing Spatial-first scanning, Temporal-first scanning, and their combination. The results indicate that BD Spatial-first scanning achieves the best performance, highlighting VideoMamba's strong scalability, sensitivity to short-term action recognition, superior long-term video understanding, and compatibility with other modalities. VideoMamba$\mathbf{^\dag}$~\cite{eccv24/VideoMamba_2} explored the reverse of Spatial-first scanning, conducting experiments on direct backward scanning, Spatial Reversal, and Temporal Reversal. The results reaffirm that BD Spatial-first scanning yields the best performance. Similarly, the Mamba architecture is employed for 3D long-term modeling in the multi-object tracking paradigm. MambaTrack~\cite{mm24/MambaTrack} leveraged Mamba to model the spatio-temporal dynamics of objects and predict their future motion. DLT~\cite{mm24/DLT} employed Mamba to process multiple frames and estimate the distances of pre-detected 2D objects in these frames from a monocular camera. VMRNN~\cite{cvpr24w/VMRNN} integrated the 2D visual Mamba block with LSTM for spatiotemporal forecasting.

\subsubsection{Video Generation}
\label{sec:video_generation}
Video generation approaches focus on capturing correlations across the spatiotemporal dimension~\cite{mm24/RainMamba,iclrw24/SSMDiff,eccv24/Zigma,arxiv24/matten,arxiv24/DiM1,neurips/VFIMamba,mm24/MMD}. RainMamba~\cite{mm24/RainMamba} introduced a 3D Hilbert scanning mechanism, while SSMDiff~\cite{iclrw24/SSMDiff}, Zigma~\cite{eccv24/Zigma}, and MMD~\cite{mm24/MMD} decoupled spatial and temporal modeling. VFIMamba~\cite{neurips/VFIMamba} utilized Temporal-first scanning to enhance spatiotemporal adjacency for the frame interpolation task. 

\begin{table}[h]
\label{tab:pointcloud}
\centering
\caption{The Mamba-based methods in point cloud vision tasks. 
The abbreviations of Task here are 
CLS: Classification, DET: Detection, 
SEG: Segmentation. Tok: Tokenization.}
\resizebox{\linewidth}{!}{
\begin{tabular}{lcccccc}
\toprule
\textbf{Methods}
& \textbf{Data}
& \textbf{Task}
& \textbf{Tok.}
& \textbf{Scan Strategy}
& \textbf{Venue}
& \textbf{Code}
\\
\midrule
\textbf{PointMamba}~\cite{neurips24/PointMamba}
& 3D Point Cloud Data
& Object CLS, Part SEG
& FPS \& KNN \& Hilbert/Trans-Hilbert
& SD Raster
& NeurIPS24
& \href{https://github.com/LMD0311/PointMamba}{\color{magenta}\Checkmark}
\\
\textbf{Mamba3D}~\cite{mm24/Mamba3D}
& 3D Point Cloud Data
& Object CLS, Part SEG
& FPS \& KNN
& Raster + Channel Flipping
& MM24
& \href{https://github.com/xhanxu/Mamba3D}{\color{magenta}\Checkmark}
\\
\textbf{LCM}~\cite{neurips24/LCM}
& 3D Point Cloud Data
& Object CLS, Object DET, Part SEG
& FPS \& KNN
& SD Raster
& NeurIPS24
& \href{https://github.com/zyh16143998882/LCM}{\color{magenta}\Checkmark}
\\
\textbf{VoxelMamba}~\cite{neurips24/voxelmamba}
& 3D Point Cloud Data
& Object DET
& Voxelization
& Multi-scale BD Hilbert
& NeurIPS24
& \href{https://github.com/gwenzhang/Voxel-Mamba}{\color{magenta}\Checkmark}
\\
\textbf{MambaMOS}~\cite{mm24/MambaMOS}
& 4D Point Cloud Data
& Object SEG
& Z/Hilbert
& SD Raster
& MM24 
& \href{https://github.com/Terminal-K/MambaMOS}{\color{magenta}\Checkmark}
\\
\bottomrule
\end{tabular}
}
\end{table}

\subsection{Point Cloud}

\label{sec:point_cloud}

Point clouds represent a fundamental 3D data structure, consisting of discrete points in space, each defined by a set of 3D coordinates. These points collectively describe the geometry of objects or scenes. However, the inherent sparse and irregular distribution of point clouds poses significant challenges for 3D vision tasks. Point clouds can be used directly as raw points or transformed into regular grids through voxelization for further processing. When using raw points, they must first undergo a serialization process during tokenization. Notably, the computational complexity of 3D point cloud analysis can be reduced by grouping points or voxels. A common approach for grouping raw points during tokenization involves sampling key points using Farthest Point Sampling (FPS), followed by selecting k-nearest neighbors (KNN) for each key point.

Inspired by the linear complexity and global modeling capabilities of Mamba, several general SSM-based backbones~\cite{cvpr24/SSM-ViT,neurips24/PointMamba,point_could_mamba,pointmambaliu,mm24/Mamba3D,arxiv24/PonTramba,neurips24/LCM,neurips24/voxelmamba,3dmambaipf,3dmambacomplete,arxiv24/EventMamba,arxiv24/OverlapMamba,mm24/MambaMOS,arxiv24/MAMBA4D,arxiv24/DiM-3D} have been explored in point cloud processing. Adaptations of 1D causal Mamba operations for 3D point clouds typically focus on scanning strategies and tokenization of raw points. For example, Z-order and Hilbert curves can be employed during scanning or tokenization to preserve spatial locality. 

PointMamba~\cite{neurips24/PointMamba} applied Hilbert and Trans-Hilbert curves to the key points obtained through FPS before performing KNN during tokenization. This study demonstrated that PointMamba, leveraging a simple Mamba block, outperforms state-of-the-art Transformer-based networks. Mamba3D~\cite{mm24/Mamba3D} identified that learning sequence order in point clouds can lead to unreliable and unstable pseudo-order dependencies. To mitigate this, channel flipping was introduced for the backward SSM. LCM~\cite{neurips24/LCM} applied the Mamba architecture to masked point modeling for pre-training, and proposed a locally constrained compact point cloud model. VoxelMamba~\cite{neurips24/voxelmamba} introduced a Multi-scale BD Hilbert scanning strategy. MambaMOS~\cite{mm24/MambaMOS} employed Z-order and Hilbert curves to serialize 4D points and designed a new embedding layer to better capture spatial and temporal information. 


\begin{table}[h!]
\label{tab:medicalVLP}
\centering
\caption{The Mamba-based methods in Multi-modal Vision Tasks. 
The abbreviations of Task here are 
DET: Detection, 
SEG: Segmentation,
SU: Scene Understanding,
VQA: Visual Question Answering. 
The abbreviations of Block here are VSS: Vision State Space Module, *: modification. }
\resizebox{\linewidth}{!}{
\begin{tabular}{lccccccc}
\toprule
\multirow{2}{*}{\textbf{Methods}} 
& \multirow{2}{*}{\textbf{Data}}
& \multirow{2}{*}{\textbf{Task}}
& \multicolumn{3}{c}{\textbf{Fusion Strategy}}                   
& \multirow{2}{*}{\textbf{Venue}}
& \multirow{2}{*}{\textbf{Code}}
\\ \cmidrule{4-6} 
                  
&  
&
& \textbf{Paradigm}
& \textbf{Mamba Location}
& \textbf{Mamba Design}
&
& 
\\
\midrule
\rowcolor{dino}
\multicolumn{8}{c}{\textbf{Homogeneous Multi-modal}} \\
\midrule
\textbf{MGMF}~\cite{tgrs24/MGMF}
& RGB Images \& Infrared Images
& Object DET
& Multi-branch
& Fusion Module
& M
& TGRS24
& 
\\
\textbf{Sigma}~\cite{wacv25/sigma}
& RGB Images \& Thermal/Depth Images
& Semantic SEG
& Multi-branch
& Backbone/Fusion Module/Decoder
& VSS/M/U/VSS*
& WACV25
& \href{https://github.com/zifuwan/Sigma}{\color{magenta}\Checkmark} \\
\midrule
\rowcolor{dino}
\multicolumn{8}{c}{\textbf{Heterogeneous Multi-modal}} \\
\midrule
\textbf{InstructGIE}~\cite{eccv24/InstructGIE}
& Natural Images \& Text
& Image Editing
& Diffusion
& Vision Encoder
& VSS
& ECCV24
& \href{https://github.com/cr8br0ze/InstructGIE-Code}{\color{magenta}\Checkmark}
\\
\textbf{Meteor}~\cite{neurips24/Meteor}
& Natural Images \& Text
& \shortstack{VQA, Reasoning, SU}
& MLLM
& LLM
& Mamba
& NeurIPS24 
& \href{https://github.com/ByungKwanLee/Meteor}{\color{magenta}\Checkmark}
\\
\textbf{RoboMamba}~\cite{neurips24/RoboMamba}
& Natural Images \& Text
& Robot Reasoning and Manipulation
& MLLM
& LLM
& Mamba
& NeurIPS24 
& \href{https://github.com/lmzpai/roboMamba}{\color{magenta}\Checkmark}
\\
\textbf{ReMamber}~\cite{eccv24/remamber}
& Natural Images \& Text
& Referring Image SEG
& Uni-branch
& Backbone
& U
& ECCV24
& \href{https://github.com/yyh-rain-song/ReMamber}{\color{magenta}\Checkmark}
\\
\textbf{RSCaMa}~\cite{grsl24/RSCaMa}
& Remote Sensing Images \& Text
& Change Captioning 
& Multi-branch
& Fusion Module
& M/U
& GRSL24
& \href{https://github.com/Chen-Yang-Liu/RSCaMa}{\color{magenta}\Checkmark}
\\
\textbf{CoupledMamba}~\cite{neurips24/Coupled_Mamba}
& Audio \& Video \& Text
& Sentiment Analysis
& Uni-branch
& Backbone
& M
& NeurIPS
&
\\
\textbf{MotionMamba}~\cite{eccv24/MotionMamba}
& Motion Sequence \& Text 
& Motion Generation
& Diffusion
& Backbone
& CA
& ECCV24
& \href{https://steve-zeyu-zhang.github.io/MotionMamba/}{\color{magenta}\Checkmark}
\\
\textbf{MambaTalk}~\cite{neurips24/mambatalk}
& Audio \& Text \& Motion
& Motion Generation
& VQ-VAE
& Fusion Module
& CA/U
& NeurIPS24
& 
\\
\textbf{MambaGesture}~\cite{mm24/MambaGesture}
& Audio \& Text \& Style \& Emotion
& Motion Generation
& Diffusion
& Backbone
& CA/A
& MM24
& 
\\
\bottomrule
\end{tabular}
}
\end{table}

\subsection{Multi-modal}
\label{sec:mm}
Multi-modal tasks have become increasingly pivotal in the domain of computer vision, enabling the enrichment of visual understanding through the integration of diverse information sources. The fundamental objective of multi-modal tasks is to learn valuable potential feature representations from a multitude of modalities, such as textual captions and visual images, RGB images with supplementary components like depth or thermal images, and various forms of medical imaging data. Nevertheless, a significant challenge in attaining multi-modal objectives lies in the effective capture and amalgamation of the information across diverse modalities. 

Early multi-modal approaches employ the \textbf{Multi-branch} paradigm, where each branch processes distinct modalities in parallel, followed by the fusion of features to create a unified representation. With the advent of Transformer architectures, a new \textbf{Uni-branch} paradigm has emerged, which concatenates various modalities and processes them within a single branch, simplifying the integration process and achieving greater efficiency. To harness the potential of large language models (LLMs), the Multi-modal Large Language Model (\textbf{MLLM}) paradigm utilizes LLMs as task decoders, facilitating enhanced multi-modal understanding. Additionally, established paradigms such as \textbf{Diffusion}~\cite{neurips20/DDPM} and \textbf{VQ-VAE}~\cite{neurips17/VQ-VAE} are widely adopted for multi-modal generation tasks, leveraging their robust generative capabilities.

Recently, Mamba has been adapted for a variety of multi-modal tasks due to its computational efficiency and flexibility in feature fusion, demonstrating impressive performance across different applications. Firstly, various modules within existing paradigms can be replaced with Mamba-based architectures to leverage its global receptive capabilities and linear computational complexity, such as backbones and vision encoders. Notably, the LLM component of MLLM paradigms can be substituted with Mamba to capitalize on its advantages. Secondly, leveraging Mamba's selective SSM mechanism, novel Mamba designs can enhance feature fusion, enabling flexible interactions between features from different modalities. Some fusion approaches employ distinct SSM mechanisms for each modality, integrating information from other modalities (individual, combined, or differences between modalities) into the SSM framework (\textbf{M}). Some fusion approaches concatenate information from different modalities, leveraging the SSM mechanism to fuse them through advanced scanning techniques (\textbf{U}). Additionally, mechanisms such as Cross Attention (\textbf{CA}) and Attention (\textbf{A}) can be incorporated into the Mamba-based architecture to facilitate effective feature fusion.

According to the modal correlation of the input data, the adaptation of Mamba to multi-modal applications can be classified into two categories: homogeneous multi-modal ($\S~\ref{Homogeneous}$) and heterogeneous multi-modal ($\S~\ref{Heterogeneous}$) applications. 

\subsubsection{Homogeneous Multi-modal}
\label{Homogeneous}

Homogeneous multi-modal applications refer to the tasks where the input data consists of multiple modalities sharing similar data type, including MRI-CT registration~\cite{mambamorph,arxiv24/VMambaMorph}, medical image fusion~\cite{mambadfuse,fusionmambaXie}, medical image generation~\cite{arxiv24/I2I-Mamba}, remote sensing images pansharpening~\cite{panmamba,fusionmambaPeng,arxiv24/LE-Mamba}, infrared-visible image fusion~\cite{mambadfuse,fusionmambaXie,arxiv24/S4Fusion}, RGB-infrared image object detection~\cite{fusion-mamba,arxiv24/CFMW,mambadfuse,tgrs24/MGMF} and RGB-thermal/depth image semantic segmentation~\cite{wacv25/sigma}.

For RGB-infrared object detection, the multi-branch paradigm MGMF~\cite{tgrs24/MGMF} introduced the Mamba-based fusion module, while utilizing the common feature extraction backbone and detection head. In particular, this fusion module maps the summed features from the two modalities into a hidden state space and employs gating mechanisms to facilitate state transitions, enabling deep cross-modal fusion of the representations. Sigma~\cite{wacv25/sigma} is another multi-branch paradigm designed for multi-modal semantic segmentation, employing a fully Mamba-based architecture across all components. Specifically, VMamba~\cite{neurips24/vmamba} blocks are used in the feature extraction backbone. For the fusion module, cross-modal multiplication and feature concatenation mechanisms are integrated within the SSM. Furthermore, channel attention is incorporated into the SSM to enhance the decoder's capability in modeling channel-wise interactions.

\subsubsection{Heterogeneous Multi-modal}
\label{Heterogeneous}
Heterogeneous multi-modal applications refer to tasks where the input data consists of multiple modalities with different data types~\cite{eccv24/InstructGIE,vlmamba,neurips24/Meteor,cobra,neurips24/RoboMamba,eccv24/remamber,arxiv24/clipmamba,survmamba,grsl24/RSCaMa,spikemba,arxiv24/BroadMamba,neurips24/Coupled_Mamba,prcv24/Mamba-FETrack,eccv24/MotionMamba,mm24/MambaGesture,tmmamba,isbi25/R2Gen-Mamba}. 

In the field of text-driven generation, Mamba architectures are utilized for visual processing. To enhance the generalization capabilities of image editing approaches, InstructGIE~\cite{eccv24/InstructGIE} reformed the conditional latent diffusion model by integrating a vision encoder based on VMamba~\cite{neurips24/vmamba}, and used the VMamba layer to inject the processed visual condition information into the frozen Stable Diffusion \cite{cvpr22/stablediff} model. In the realm of MLLMs, Meteor~\cite{neurips24/Meteor} introduced multifaceted rationales and applied the Mamba architecture to embed these lengthy rationales to enhance the understanding and answering capabilities of MLLMs. RoboMamba~\cite{neurips24/RoboMamba} aligned the vision encoder with an efficient Mamba language model to enable cost-effective robot manipulation capabilities. ReMamber~\cite{eccv24/remamber} introduced a novel referring image segmentation framework containing multiple Mamba twister blocks. Each block comprises VMamba~\cite{neurips24/vmamba} layers to extract the vision feature and a twister layer to integrate the textual information into the visual modality. The Twister layer first constructs a hybrid feature cube and subsequently performs channel-wise and spatial scanning for feature fusion. RSCaMa \cite{grsl24/RSCaMa} employed multiple CaMa layers between the backbone and the language decoder to facilitate efficient joint spatial-temporal modeling for remote sensing image change captioning. Specifically, the CaMa layer enhances change awareness by multiplying the differencing features with the output of SSMs. It further employs a Temporal-first scanning strategy to facilitate effective temporal modeling. 

CoupledMamba~\cite{neurips24/Coupled_Mamba} proposed coupling state chains of inter-modalities while maintaining the independence of intra-modality state processes. It also derived the global convolution kernel to enable parallelism for the coupled state transition scheme. 
MotionMamba~\cite{eccv24/MotionMamba} incorporated two modules based on the vanilla Mamba into a diffusion-based generative system for text-to-motion synthesis, namely hierarchical temporal Mamba and bidirectional spatial Mamba. MambaGesture~\cite{mm24/MambaGesture} integrated Mamba architecture with the attention mechanism to form the basic block for the diffusion model in the co-speech gesture generation task.

\section{Challenges and Future Directions}
\label{sec:challenges_and_future}
In this section, we identify current challenges and outline promising directions, to enhance Mamba's capabilities and broaden its applications in computer vision.

\subsection{Scalability}

\subsubsection{Model Size}  
In the current era of large foundation models, the scalability of deep neural networks is pivotal for advancing general artificial intelligence. Mamba is emerging as a promising foundational architecture, offering an alternative to prevalent Transformers with its ability to scale linearly with sequence length. However, while existing techniques aimed at reducing the computational complexity of Transformers often compromise on model expressivity or adaptability, Mamba also encounters challenges when scaled to large-sized networks.

Firstly, the Mamba architecture faces stability issues when scaled to large network configurations \cite{arxiv24/mamba360}. The underlying causes of Mamba's instability remain inadequately understood. This instability often manifests as vanishing or exploding gradients \cite{arxiv24/simba}, leading to performance degradation or even training collapse. To address the stability issues, some studies utilize stable reparameterization to balance gradient scales in state space models \cite{icml24/stablessm,neurips24/complex,icml24/universality}. Some works utilize learnable Fourier Transforms to adjust the eigenvalues of the evolution matrix, ensuring they are negative real numbers, which helps maintain stability \cite{arxiv24/simba}. Mamba-2 incorporates an additional normalization layer in the block before the final output projection, mitigating instability issues encountered in large models~\cite{arxiv24/mamba-2}. Despite these efforts, most visual Mamba models are limited to base or smaller scales, which may restrict their performance.

Secondly, visual Mamba models, when controlled for similar parameters or FLOPs, are generally outperformed by cutting-edge CNNs, advanced visual Transformers, and hybrid architectures that integrate multiple model types. This performance gap also hinders the scaling of visual Mamba models and underscores the necessity to enhance their performance. Integrating architectures or techniques proven effective in CNNs and visual Transformers could help bridge the performance gap. Besides, while there may be some compromise in performance, the non-hierarchical architectures comprising identical blocks facilitate optimization for hardware acceleration. This aspect is crucial for scaling to large models and enhancing computational efficiency. In addition, self-supervised pre-training plays a critical role in enhancing the model performance. For instance, ARM \cite{arxiv24/ARM} demonstrated that the autoregressive pre-training can significantly improve the performance of the Mamba architecture and address the instability issues, efficiently unlocking its potential for scaling to very large model sizes. However, the self-supervised pre-training strategies for Mamba models are still nascent. 

More recently, Mamba-2~\cite{arxiv24/mamba-2} outlined several potential directions for further enhancing its performance. Given that SSD enhances hardware efficiency by limiting expressivity through the use of restricted structured matrices, future work could explore refining structured matrix algorithms toward more generalized diagonal SSMs to balance efficiency and expressivity~\cite{icml24/illusion}. Mamba-2 further suggests that advancements from a matrix mixer perspective could broaden the applications of sequence models, including the development of principled non-causal Mamba variants and methods to bridge the gap between softmax attention and sub-quadratic models by analyzing their matrix transformation structures.

Future research should explore optimizing these aspects to harness the full capabilities of large network configurations.

\subsubsection{Data}

The Mamba architecture theoretically reduces computational complexity, enhancing its efficiency in long sequence processing. Moreover, unlike many sequence models that do not benefit from extended contexts \cite{icml23/gsm-ic}, Mamba can selectively disregard irrelevant information, yielding consistent performance improvements as context length increases. The efficiency and efficacy of Mamba architecture bring significant opportunities for data scalability. 

Firstly, the Mamba architecture offers significant potential for effectively handling high-dimensional data by treating each sample as a long sequence. This capability is particularly beneficial for processing high-resolution and multi-dimensional data including high-resolution remote sensing images, whole slice pathology images, highspectral and multispectral images, 3D medical images, and point clouds. Secondly, its efficiency and prowess in long sequence modeling not only facilitate scaling to large-scale datasets, but also enable effective analysis of the temporal or spatial relationships among samples. The typical datasets include multi-temporal remote sensing images, where changes over time provide essential information, and long-term video frames, where continuous activities must be analyzed over extended periods. 

Furthermore, given the parameter efficiency of the Mamba models and their inherent local inductive bias, which typically mitigate overfitting, Mamba models have considerable potential for achieving optimal performance without relying on large-scale datasets. This opens up opportunities for developing small yet effective Mamba-based models advantageous in resource-constrained environments, a pathway that is underexplored in current research. 

Future research should concentrate on scaling Mamba models for processing visual data with higher dimensions and larger scales. Fully exploiting the visual data by designing small models that maintain high performance is also a promising future direction.

\subsubsection{Hardware-aware}
Previous SSMs such as S4 are time- and input-invariant, making it feasible to unroll the recurrent representation into the convolutional representation for efficient parallel computation during training. However, the selection mechanism in Mamba makes the model become time- and input-dependent, making it infeasible to train the model under the convolutional representation. To overcome this limitation of the selective SSMs (Mamba) and make them efficient on modern hardware, hardware-aware algorithm has been designed to accelerate the training under the recurrent mode. 

Mamba-1 uses kernel fusion and recomputation to make the computation of the parallel SSM scan fast and memory-efficient. Specifically, the discretization step, the scan, and the multiplication with the projection matrix are fused into one kernel. Then the SSM parameters are loaded from slow HBM to fast SRAM for calculating the fused kernel in SRAM. Subsequently, the outputs are written back to HBM. In addition, Mamba-1 chooses to recompute the intermediate states for the backpropagation since the recomputation is faster than storing them and reading them from HBM. These two strategies accelerate the scan operation and reduce the GPU requirement. However, according to research findings \cite{arxiv24/mamba_survey,arxiv24/mamba-2}, GPU consumption for the Mamba model does not consistently demonstrate a reduction compared to the Transformer model, highlighting the need for optimized, hardware-aware Mamba algorithms tailored for vision tasks. 

Mamba-2 \cite{arxiv24/mamba-2} establishes theoretical connections between structured SSMs and variants of attention by reformulating different approaches to computing SSMs as various matrix multiplication algorithms on structured matrices and generalizing the linear attention \cite{icml20/linear_att} to structured masked attention. Based on this theoretical analysis, Mamba-2 introduces an efficient SSD algorithm for SSM computation by employing block decompositions of particular semiseparable matrices and modifies the Mamba block to be tensor parallelism friendly. This framework allows Mamba-2 to leverage system optimizations developed for Transformers, including tensor parallelism and sequence parallelism, facilitating the parallel training of large models using multiple GPUs. Moreover, unlike Transformers, Mamba-2 can be efficiently trained with variable sequence lengths. Given that semiseparable matrices are extensively studied in scientific computing, incorporating these techniques presents promising opportunities for further advancing SSMs~\cite{arxiv24/mamba-2}. Besides, the adaptation of Mamba-2 to vision tasks requires further investigation. 

Future research should focus on developing hardware-aware approaches to enhance the computational efficiency of visual Mamba models. On the one hand, hardware-aware approaches that reduce computational overhead are essential for scaling to larger hardware configurations. On the other hand, from a broader perspective, they are crucial for the development of green artificial intelligence technologies, which attract more and more attention from the community due to the need for energy consumption reduction to protect the environment.

\subsubsection{Task}
The Mamba architecture, noted for its efficiency and proficiency in long sequence modeling, enables applications across a broader spectrum of tasks compared to CNNs and Transformers. 

Firstly, as previously mentioned, the Mamba architecture offers broader possibilities for handling high-dimensional data and large-scale datasets. 

Secondly, the Mamba model's proficiency in handling extended sequences significantly enhances its applicability in multi-modal learning. Similar to the Transformer, which effectively models both natural languages and images within a unified framework, the long sequence modeling capability of the Mamba architecture allows it to involve more modalities such as time series and extended textual content. Besides, like the Transformer, Mamba can adeptly handle concatenated independent sequences, processing them either as a whole or as distinct entities by resetting their states at boundaries. Furthermore, Mamba offers computational efficiency and time-scale robustness, the latter of which is particularly advantageous for addressing issues with missing modalities in data streams.

Thirdly, the high efficiency of the Mamba architecture enables the development of high-speed and highly efficient applications. This includes interactive systems, autonomous driving, and surveillance, where real-time processing and reduced computational overhead are critical.

Future research should focus on effectively leveraging the Mamba architecture to enhance performance in existing tasks and to explore new applications. 

\subsection{Causality}

In visual Transformer-based architectures, the attention matrix explicitly captures dependencies between each pair of visual tokens, enabling predictions at any location to have a global receptive field. However, in Mamba-based architectures, the S6 blocks make predictions for each token based solely on the hidden state $h_{t-1}$ and the current input $x_t$. The $h_{t-1}$ only incorporates information from previously scanned tokens and disregards information from other tokens. Consequently, the vanilla Mamba is a causal system where predictions rely solely on the current and previous inputs. 
Although the causality is naturally suitable for autoregressive models in sequential vision tasks~\cite{arxiv24/ARM}, it contradicts the nature of spatial images, resulting in a fundamental mismatch between Mamba-based architectures and common visual tasks. To address this issue, existing works have proposed to integrate multiple scanning techniques to ensure a global receptive field. While these strategies partially mitigate the undesirable causality of Mamba in vision tasks, they still encounter challenges or introduce new drawbacks that require resolution.

Firstly, while the utilization of multiple scanning techniques helps establish a global receptive field, this approach does not effectively address the underlying causality issue. Tokens at different locations are still treated inconsistently due to their varying positions within the sequences, inevitably leading to a prioritization of more recently scanned tokens. Consequently, this strategy struggles to preserve spatial relationships and inadvertently introduces undesired bias for vision tasks as a result of the scanning order. In contrast, the self-attention mechanism in Transformers consistently considers all tokens to capture dependencies, showcasing its superiority to Mamba. 

Secondly, the incorporation of multiple scanning strategies leads to redundancy and sub-optimal performance. To enhance Mamba's perception at different locations, numerous approaches incorporate different scanning techniques including different scanning directions, axes, continuity, and sampling. While incorporating more scanning strategies may increase the information captured in specific locations, it introduces a significant amount of redundant information within these scanning sequences. This redundancy can impact model efficiency and hinder effective knowledge extraction for vision tasks.

Consequently, designing an appropriate approach for Mamba to consistently and elegantly consider tokens at different locations remains a critical challenge in visual data processing.

\subsection{In-context Learning}
Transformer models exhibit in-context learning (ICL) capabilities after large-scale pre-training, enabling them to learn new tasks with a few demonstrations without further explicit training or fine-tuning. In the dynamic landscape of deep learning, ICL methodologies have evolved to address increasingly complex tasks across NLP, CV, and multi-modal domains. These methodological advancements are pivotal for pushing the limits of existing deep learning frameworks. 

Currently, Transformer models are the primary large models possessing ICL capabilities. Mamba's efficiency and long sequence modeling ability make it a promising alternative to Transformers for ICL. Furthermore, the inherent causality of the Mamba model eliminates the need for position embeddings, which are typically a constraint in extending the ICL capabilities of Transformers. In \cite{icml24/ICL2}, the authors investigated the ICL capabilities of Mamba compared to Transformers, focusing on regression ICL tasks and small-scale models. The results demonstrated that Mamba is capable of ICL and performs comparably to Transformers in standard regression ICL tasks. Additionally, Mamba outperforms Transformers in tasks such as sparse parity learning. However, Mamba struggles with tasks involving non-standard retrieval functionality, which Transformers can handle with ease. This limitation may stem from the memory-intensive nature of these tasks, as Mamba prioritizes efficiency over contextual expressivity by utilizing a shorter latent size~\cite{icml24/repeat_after_me}. Mamba-2 demonstrates notable improvements over Mamba-1 in some of these tasks, though further investigation is needed to fully understand its performance~\cite{arxiv24/mamba-2}. Research detailed in \cite{automl24/ICL1} further demonstrated Mamba's capabilities on more complex NLP in-context tasks and showed that Mamba's ICL efficacy scales positively with the number of in-context examples.

With its proficient in-context modeling capabilities and adeptness at capturing long-range dependencies, the Mamba model shows promising potential for deeper semantic understanding and enhanced performance in ICL applications.

\subsection{Trustworthiness}
\subsubsection{Interpretability}
Some studies have provided experimental evidence to elucidate the mechanisms underlying the Mamba model in NLP, focusing on its in-context learning capabilities \cite{automl24/ICL1,icml24/ICL2}, factual recall capabilities \cite{colm24/factual}, and next-token prediction capabilities~\cite{icml24/next-token}. Additionally, other works \cite{arxiv24/theo_mamba} have laid theoretical foundations for Mamba's applications in NLP. Despite these advancements, explaining why Mamba performs effectively on visual tasks remains challenging~\cite{neurips24/MambaLRP,arxiv24/att_mamba}. In \cite{arxiv24/att_mamba}, the authors reinterpret Mamba layers as self-attention mechanisms, thereby shedding light on its functionality both empirically and theoretically. However, the distinct learning characteristics of visual Mamba and its parallels with other foundational architectures, such as RNNs, CNNs, and ViTs, still demand deeper interpretation.

\subsubsection{Generalization}
According to findings in \cite{mm24/dgmamba,neurips24/START}, the hidden states in Mamba are likely to accumulate or even amplify domain-specific information, which can adversely affect its generalization performance. Moreover, the 1D scanning strategies inherent to the model may inadvertently capture domain-specific biases, and current scanning techniques often fail to address the need for domain-agnostic information processing. 
Addressing these challenges to better harness Mamba's strengths while improving its generalization remains a critical area for future research.

\subsubsection{Safety}
Research presented in \cite{arxiv24/robust_mamba} demonstrated VMamba's \cite{neurips24/vmamba} strengths in adversarial resilience and general robustness. However, it also identified limitations in scalability when dealing with these tasks. The study includes white-box attacks on VMamba to examine the behavior of its novel components under adversarial conditions. The findings indicated that while parameter $\Delta$ exhibits robustness, parameters $\mathbf{B}$ and $\mathbf{C}$ are susceptible to attacks. This differential vulnerability among parameters contributes to VMamba's scalability challenges in maintaining robustness. Furthermore, the results revealed that VMamba exhibits particular sensitivity to interruptions in the continuity of its scanning trajectory and the integrity of spatial information. Enhancing the safety of visual Mamba remains an unresolved challenge.

\section{Conclusion}
Mamba has rapidly emerged as a transformative long sequence modeling architecture, renowned for its exceptional performance and efficient computational implementation. As it continues to gain significant attraction in the field of computer vision, this paper offers a comprehensive review of visual Mamba approaches. We begin with an in-depth overview of the Mamba architecture, progressing to detailed examinations of representative visual Mamba backbone networks and their extensive applications. These applications are systematically categorized by differing modalities, including image, video, point cloud, and multi-modal data. Finally, we analyze the challenges and delineate future directions for visual Mamba, providing valuable outlooks that may influence ongoing and future developments in this dynamically evolving field.

\begin{acks}
This work was supported by the Hong Kong Innovation and Technology Fund (Project No. MHP/002/22), Project of Hetao Shenzhen-Hong Kong Science and Technology Innovation Cooperation Zone (HZQB-KCZYB-2020083) and the Research Grants Council of the Hong Kong (Project Reference Number: T45-401/22-N). 
\end{acks}

\bibliographystyle{ACM-Reference-Format}
\bibliography{mamba}










\end{document}